\documentclass{article}

     \usepackage[preprint]{neurips_2023}

\usepackage[utf8]{inputenc} 
\usepackage[T1]{fontenc}    
\usepackage{hyperref}       
\usepackage{url}            
\usepackage{booktabs}       
\usepackage{amsfonts}       
\usepackage{nicefrac}       
\usepackage{microtype}      
\usepackage{xcolor}         

\usepackage{graphicx}

\usepackage{amsmath}
\usepackage{amssymb}
\usepackage{mathtools}
\usepackage{amsthm}

\usepackage{comment}
\usepackage{multirow}

\usepackage{tikz}
\usetikzlibrary{patterns}

\usepackage{subcaption}

\def\argmax{\operatornamewithlimits{argmax}}
\def\argmin{\operatornamewithlimits{argmin}}

\usepackage{siunitx}

\title{DeepEMD: A Transformer-based Fast Estimation of the Earth Mover's Distance}

\author{%
  Atul Kumar Sinha\\
  University of Geneva\\
  \texttt{atul.sinha@unige.ch} \\
  \And
  François Fleuret \\
  University of Geneva\\
  \texttt{francois.fleuret@unige.ch} 
}

\begin{document}

\maketitle

\begin{abstract}
  The Earth Mover's Distance (EMD) is the measure of choice between point clouds. However the computational cost to compute it makes it prohibitive as a training loss, and the standard approach is to use a surrogate such as the Chamfer distance. 
  We propose an attention-based model to compute an accurate approximation of the EMD that can be used as a training loss for generative models. To get the necessary accurate estimation of the gradients we train our model to explicitly compute the matching between point clouds instead of EMD itself. We cast this new objective as the estimation of an attention matrix that approximates the ground truth matching matrix. 
  Experiments show that this model provides an accurate estimate of the EMD and its gradient with a wall clock speed-up of more than two orders of magnitude with respect to the exact Hungarian matching algorithm and one order of magnitude with respect to the standard approximate Sinkhorn algorithm, allowing in particular to train a point cloud VAE with the EMD itself. 
  Extensive evaluation show the remarkable behaviour of this model when operating out-of-distribution, a key requirement for a distance surrogate. Finally, the model generalizes very well to point clouds during inference several times larger than during training. \footnote{Project page: \url {https://github.com/atulkumarin/DeepEMD}}

  
%
\end{abstract}

\section{Introduction}

\label{sec:intro}

%
%
\begin{figure}[t]
\hspace*{-2em}
\begin{minipage}[t]{0.5\linewidth}
\begin{tabular}{ccc}%
\includegraphics[width=0.33\linewidth,trim=50 45 35 0, clip]{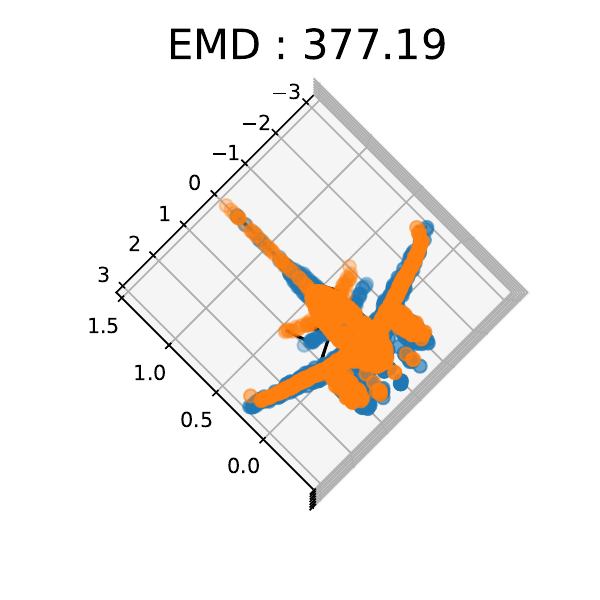}&
\includegraphics[width=0.33\linewidth,trim=50 45 35 0, clip]{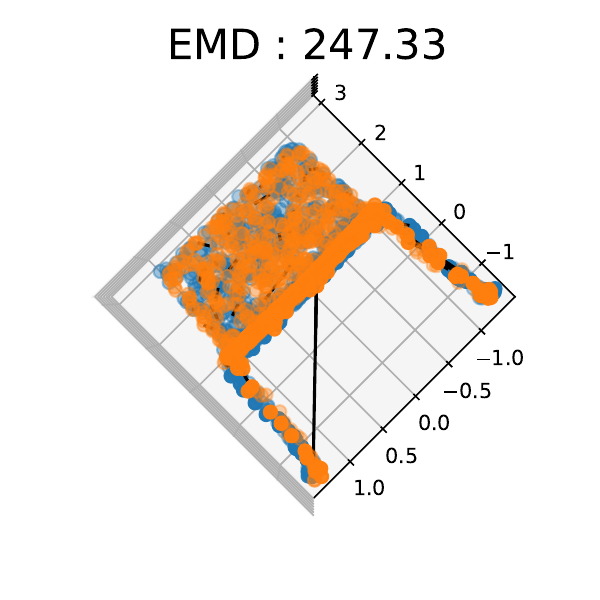}&
\includegraphics[width=0.33\linewidth,trim=50 45 35 0, clip]{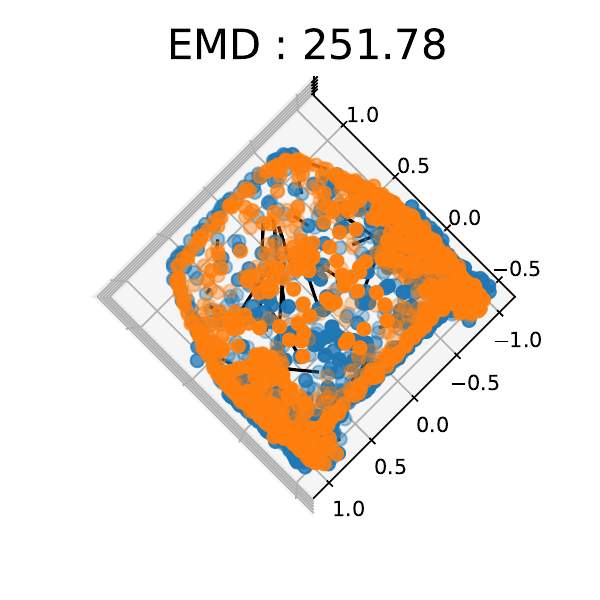}\\
\multicolumn{3}{c}{Trained with Chamfer distance}\\[2ex]
\includegraphics[width=0.33\linewidth,trim=50 45 35 0, clip]{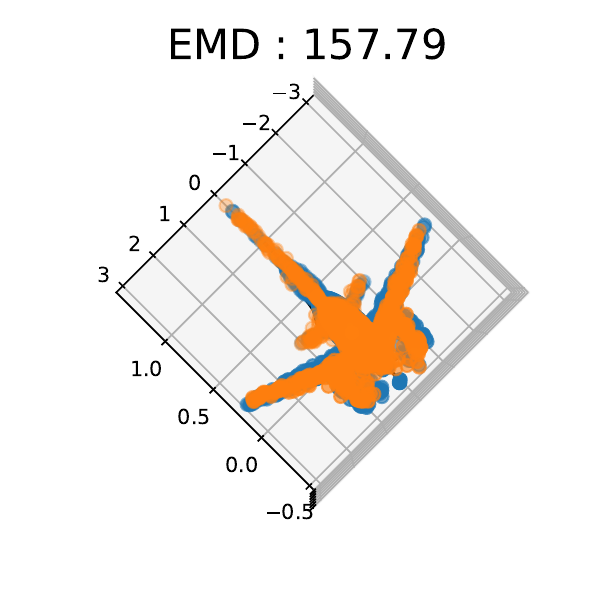}&
\includegraphics[width=0.33\linewidth,trim=50 45 35 0, clip]{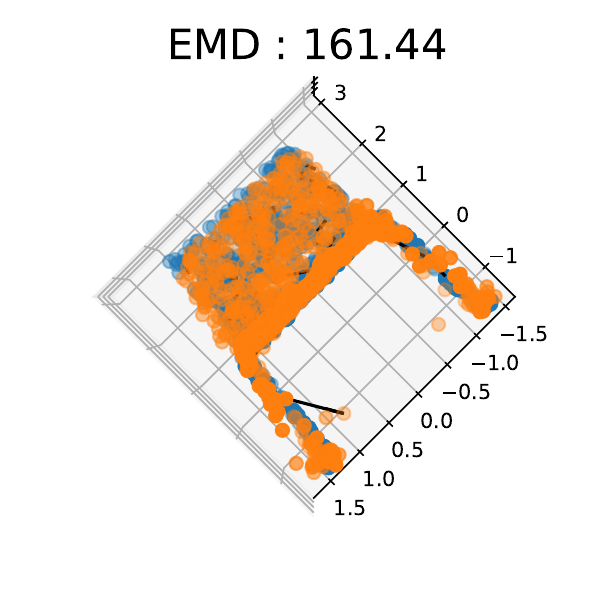}&
\includegraphics[width=0.33\linewidth,trim=50 45 35 0, clip]{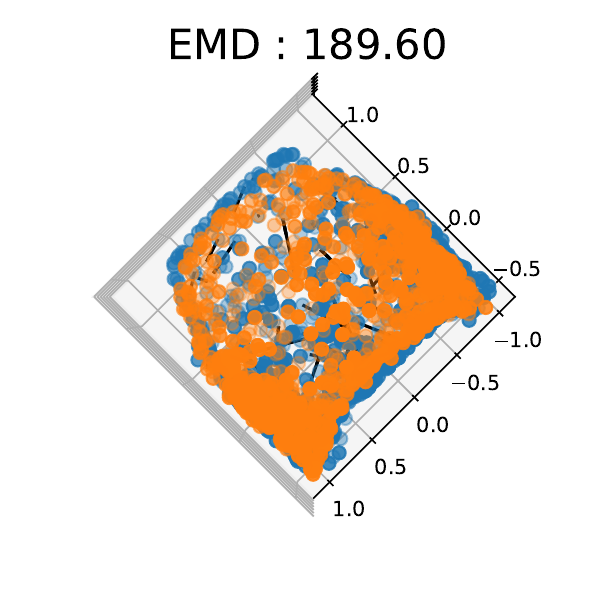} \\
\multicolumn{3}{c}{Trained with DeepEMD (ours)}
\end{tabular}
\caption{Example point clouds (blue) and their VAE reconstructions (orange) when trained with different reconstruction losses. Training with DeepEMD (bottom) consistently achieves lower reconstruction error (EMD, shown on top of each example) than with the standard Chamfer distance (top).}
\label{fig:val_recon_intro}
\end{minipage}%
    \hfill%
\begin{minipage}[t]{0.45\linewidth}
\centering
     \begin{subfigure}{\columnwidth}
         \centering
         \includegraphics[width=\textwidth,trim={25 0 22.8cm 0},clip]{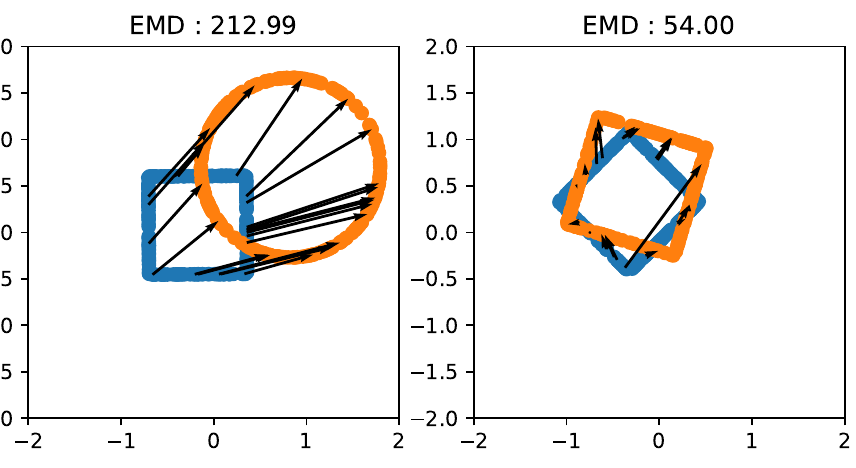}\\
         \label{fig:example_pcs_emd}
     \end{subfigure}
     
     \hfill
     \begin{subfigure}{\columnwidth}
         \centering
         \includegraphics[width=\textwidth,trim={25 0 22.8cm 0},clip]{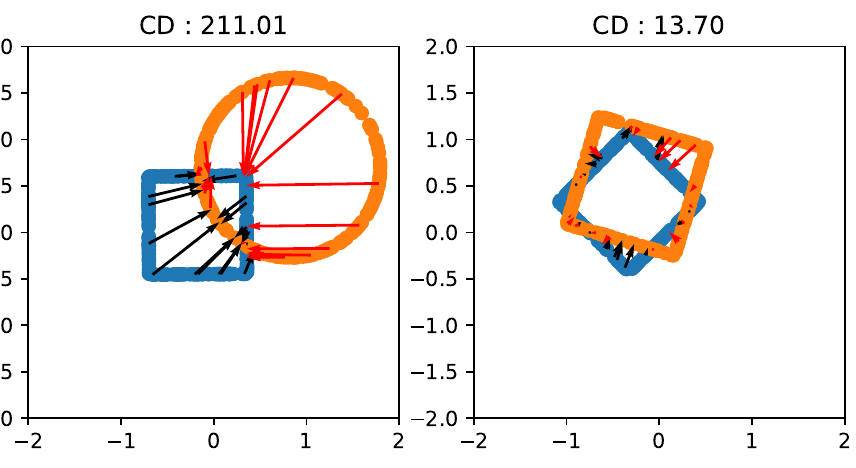}
         \label{fig:example_pcs_cd}
     \end{subfigure}
        \caption{Example pairs of point clouds. The true earth mover's distance (EMD, top) and Chamfer distance (CD, bottom) are shown above each example. Arrows indicate the matching between the two point clouds under their respective metrics.}
        \label{fig:example_pcs}
\end{minipage}
\end{figure}
%

The \emph{earth mover's distance} (EMD), also known as \emph{Wasserstein distance} is a distance between distributions that is defined as the minimum total of mass-time-distance displacement needed to transform one distribution to the other.  In the case of uniform distributions over a finite number of points, it turns into a distance between point clouds that corresponds to finding the one-to-one matching that minimizes the sum of the distances between pairs of matched points. Since there is no inherent ordering in point cloud data, computing the EMD between two point clouds involves finding a matching based on the euclidean distance between points. The matching is constrained to be bipartite so that one point cloud is completely transformed to the other, without any fractional assignment, and the transport cost is minimal.

The EMD is the most commonly used distance metric on point clouds, and is extremely useful in many different contexts. In particular as we will see for both assessing the performance of, and for training variational autoencoders, since the generated point cloud should get as close as possible to the target in terms of displacement. It can also be interpreted as the distance between two distributions computed with a finite number of samples and reflects the notion of nearness properly, does not have quantization/binning and non-overlapping support problems of most other metrics, e.g., $f$-divergences, total variation distance, etc.

The EMD between point clouds can be computed exactly, but it is extremely expensive computationally. The standard method is the Hungarian matching \cite{kuhn1955hungarian} algorithm whose complexity is $O(N^3)$ where $N$ is the number of points. Due to this computational cost, training deep generative models for point clouds is not done with this metric, even though it is a metric of choice for performance evaluation. The standard approach uses the \emph{Chamfer distance} (CD) as surrogate. This metric can be computed in $O(N^2)$ time complexity but relaxes the one-to-one matching, which may create pathological situations.

We propose a deep architecture that takes as input two point clouds encoded as sets of geometric coordinate tuples, and computes an accurate estimate of the EMD. We show that the most efficient approach, in particular if that estimate is used as a loss for a training process, is to estimate the matching matrix itself. Since the EMD is the sum of the distance between matched points, this approach provides a very accurate estimate of the gradient with respect to the point coordinates. Training a deep variational autoencoder with our model instead of an exact computation is up to $\times 100$ faster (see Fig. \ref{fig:eval_time}), and the resulting model performs far better than one trained with the usual Chamfer surrogate (see \S~\ref{sec:genModels} and Fig. \ref{fig:recon_emd}). 

The key contributions of this paper can be summarized as:
\begin{itemize}
    \setlength\itemsep{0.5em}
    \item We propose DeepEMD which approximates the EMD between point clouds in $O(N^2)$ time complexity vs $O(N^3)$ of the hungarian algorithm (\S~\ref{sec:method}).
    \item We propose to cast the prediction of a bipartite matching as an attention matrix from which we get an accurate estimate
    of the EMD and its gradient (\S~\ref{sec:deepEMD}).
    \item We show that DeepEMD  generalizes well to unseen data distributions (\S~\ref{sec:results}), and can be used for evaluation of generative models. It provides accurate estimates of the gradients of the distance and demonstrate that DeepEMD can be used as a surrogate reconstruction loss for training deep generative models of point clouds (\S~\ref{sec:genModels}).
    \item We show that DeepEMD achieves about $40\times$ speed-up over Sinkhorn algorithm, achieving equal or better performance for various metrics (\S~\ref{sec:results}).
\end{itemize}

\section{Related Work}
\label{sec:related}

The two commonly used distance metrics for point clouds in literature are Earth Mover's Distance (EMD) and Chamfer Distance (CD). Consider two point clouds $X = \{x_i\}_{i=1}^N$ and $Y = \{y_j\}_{j=1}^N$, where $x_i, y_j \in \mathbb{R}^d$. The EMD between the two point clouds can be computed as,
\begin{align}
\operatorname{EMD}(X, Y) &=\min _{\phi \in \mathcal{M}(X, Y)} \sum_{x \in X}\|x-\phi(x)\|_2,
\label{eq:emd}
\end{align}
where $\mathcal{M}(X,Y)$ is the set of 1-to-1 (bipartite) mappings from $X$ to $Y$. In addition to the distance, the optimal matching $\phi^*$ is also interesting for some applications.
Since directly optimizing the EMD is computationally expensive, most methods in the literature rely on CD as a proxy similarity measure or reconstruction loss. The CD can be computed as,
\begin{align}
\operatorname{CD}(X, Y)=\sum_{x \in X} \min _{y \in Y}\|x-y\|_2^2+\sum_{y \in Y} \min _{x \in X}\|x-y\|_2^2,
\end{align}
and in ${O}(N^2)$ time complexity. The CD solution leads to a non-bipartite one-to-many matching between $x \rightarrow y$ and vice versa. We can also use the $L_2$ measure with $d = \|x-y\|_2$ instead $d = \|x-y\|_2^2$ to make it comparable to EMD. of Note that the above EMD for point clouds is related to the Wasserstein-$2$ metric (see appendix \S~\ref{sec:app_ot} for details). 
%
%
The utility of EMD is limited by the $O(N^3)$ computational cost of evaluating it.  There have been several research efforts to circumvent this issue in various application settings.

This is the case for application to point clouds where $N$ is usually in the range of several thousands. \citet{kim2021setvae} trains a variational auto-encoder with CD as the reconstruction loss. EMD is still the metric of choice for evaluating point cloud generative models \cite{huang2022learning, luo2021diffusion, kim2021setvae, yang2019pointflow, shu20193d, achlioptas2018learning}. Another issue is disparity between performance measures (minimum matching distance, coverage, etc.) computed with EMD and CD, the comparisons are contradictory and often inconsistent across measures.

CD is usually insensitive to mismatched local density while EMD is dominated by global distribution and overlooks the fidelity of detailed structures \cite{wu2021density}. \citet{wu2021density} proposes a new similarity metric called Density-aware Chamfer distance (DCD) to tackle these issues. DCD is derived from CD and can also be computed in $O(N^2)$ time complexity. \citet{urbach2020dpdist} proposed Deep Point Cloud Distance (DPDist) which measures the distance between the points in one cloud and the estimated continuous surface from which the other point cloud is sampled. The surface is estimated locally by a network using the 3D modified Fisher vector representation. 

\citet{shirdhonkar2008approximate} proposed a linear time algorithm for approximating the EMD by exploiting the Hölder continuity constraint in its dual form to convert it into a simple optimization problem with an explicit solution in the wavelet domain and computed as the sum of absolute values of the weighted wavelet co-efficients of the difference histogram. However, their approach is limited to low dimensional histograms. In the optimal transport literature, several efforts have been taken towards improving the statistical and computational properties. Recently, \citet{chuang2022infoot} proposed Information Maximizing Optimal Transport (InfoOT) which is an information-theoretic extension of optimal transport based on kernel density estimation of the mutual information which introduces global structure into OT maps. The resulting solution maximizes the mutual information between domains while minimizing geometric distance and improves the capability for handling data clusters and outliers. 

Other approaches focus on regularizing the OT problem for making it smooth and strictly convex \cite{cuturi2013sinkhorn, flamary2016optimal, genevay2018learning, blondel2018smooth}. \emph{Sinkhorn distances} \cite{cuturi2013sinkhorn} smooth the classic OT problem with an entropic regularization term and can be computed through Sinkhorn’s matrix scaling algorithm at a speed that is several orders of magnitude faster than that of transport solvers. We provide more details in the appendix. Meta OT \cite{amos2022meta} proposes a meta model to predict the solution to the optimal transport problem which is then used to initialize a standard Sinkhorn solver to further refine the predicted solution. The architecture of the meta model depends on the data domain, and DeepEMD can be utilized when working with point clouds. The choice of meta model architecture is contingent upon the specific data domain, and DeepEMD demonstrates its exceptional utility in point cloud processing.



In this paper, our goal is to approximate the EMD using a deep network in a learning based paradigm where each sample represents two distributions and the target for regression is either the true metric i.e. EMD or the optimal matching $\phi^*$, or both. Existing point cloud datasets can serve as an interesting learning problem, where we can interpret each point cloud as a 2D or 3D distribution of points on a shape (manifold). It is posed as a supervised learning problem where the task is to estimate the true EMD, or the true bipartite matching, or both, between a pair of input point clouds.

\section{Method}
\label{sec:method}


\begin{figure}[ht]
\begin{minipage}[t]{0.5\linewidth}
\center
\resizebox{2in}{!}{%
\begin{tikzpicture}[scale=0.6666666666]

\node[draw,minimum width=0.75cm,minimum height=0.5cm] (u1) at (0, 0) {$u_1$};
\node at (1.5,0) {$\dots$};
\node[draw,minimum width=0.75cm,minimum height=0.5cm] (uN) at (3, 0) {$u_N$};

\node[draw,fill=blue!25,minimum width=0.75cm,minimum height=1cm] (mlpu1) at (0, 1.5) {\rotatebox{90}{mlp}};
\node at (1.5,2) {$\dots$};
\node[draw,fill=blue!25,minimum width=0.75cm,minimum height=1cm] (mlpuN) at (3, 1.5) {\rotatebox{90}{mlp}};

\node[draw,fill=blue!25,minimum height=0.7cm,minimum width=0.7cm] (plusu) at (1.5, 3) {$+$};

\node[draw,minimum width=0.75cm,minimum height=0.5cm] (eu) at (4, 3.5) {$e_u$};

\draw (u1)--(mlpu1); \draw (mlpu1.north)--(plusu);
\draw (uN)--(mlpuN); \draw (mlpuN.north)--(plusu);
\draw (plusu)--(eu);

\node[draw,minimum width=0.75cm,minimum height=0.5cm] (v1) at (5, 0) {$v_1$};
\node at (6.5,0) {$\dots$};
\node[draw,minimum width=0.75cm,minimum height=0.5cm] (vN) at (8, 0) {$v_N$};

\node[draw,fill=blue!25,minimum width=0.75cm,minimum height=1cm] (mlpv1) at (5, 1.5) {\rotatebox{90}{mlp}};
\node at (6.5,2) {$\dots$};
\node[draw,fill=blue!25,minimum width=0.75cm,minimum height=1cm] (mlpvN) at (8, 1.5) {\rotatebox{90}{mlp}};

\node[draw,fill=blue!25,minimum height=0.7cm,minimum width=0.7cm] (plusv) at (6.5, 3) {$+$};

\node[draw,minimum width=0.75cm,minimum height=0.5cm] (ev) at (4, 4.25) {$e_v$};

\draw (v1)--(mlpv1); \draw (mlpv1.north)--(plusv);
\draw (vN)--(mlpvN); \draw (mlpvN.north)--(plusv);
\draw (plusv)--(ev);

\node[draw,fill=blue!25,minimum width=0.75cm,minimum height=1cm] (mlpfinal) at (4, 6) {\rotatebox{90}{mlp}};

\draw (ev)--(mlpfinal);

\node[draw] (dhat) at (4,8) {$\hat{d}$};
\draw (mlpfinal)--(dhat);

\end{tikzpicture}
}
\caption{The MLP model (see \S~\ref{sec:baseline}) \mbox{predicts} directly an estimate $\hat{d}$ of the EMD.}\label{fig:mlp}

\end{minipage}%
\quad%
\begin{minipage}[t]{0.5\linewidth}
\center
\resizebox{2in}{!}{%
\begin{tikzpicture}[scale=0.6666666666]
\node[draw,fill=green!20,minimum width=0.75cm,minimum height=1cm] (alpha1) at (-0.8, 0.5) {$\alpha$};
\node[draw,fill=white,minimum width=0.75cm,minimum height=1cm] (u1) at (0, 0) {$u_1$};
\node at (1.1,0) {$\dots$};
\node[draw,fill=green!20,minimum width=0.75cm,minimum height=1cm] (alphaN) at (2.2, 0.5) {$\alpha$};
\node[draw,fill=white,minimum width=0.75cm,minimum height=1cm] (uN) at (3, 0) {$u_N$};

\node[draw,fill=blue!20,minimum height=0.7cm,minimum width=0.7cm] (plusu1) at (0   ,2) {$+$};
\node at (1.5,2) {$\dots$};
\node[draw,fill=blue!20,minimum height=0.7cm,minimum width=0.7cm] (plusuN) at (3   ,2) {$+$};
\draw (alpha1)--(plusu1); \draw (u1)--(plusu1);
\draw (alphaN)--(plusuN); \draw (uN)--(plusuN);

\node[draw,fill=green!20,minimum width=0.75cm,minimum height=1cm] (beta1) at (4.2, 0.5) {$\beta$};
\node[draw,fill=white,minimum width=0.75cm,minimum height=1cm] (v1) at (5, 0) {$v_1$};
\node at (6.1,0) {$\dots$};
\node[draw,fill=green!20,minimum width=0.75cm,minimum height=1cm] (betaN) at (7.2, 0.5) {$\beta$};
\node[draw,fill=white,minimum width=0.75cm,minimum height=1cm] (vN) at (8, 0) {$v_N$};

\node[draw,fill=blue!20,minimum height=0.7cm,minimum width=0.7cm] (plusv1) at (5   ,2) {$+$};
\node at (6.5,2) {$\dots$};
\node[draw,fill=blue!20,minimum height=0.7cm,minimum width=0.7cm] (plusvN) at (8   ,2) {$+$};
\draw (beta1)--(plusv1); \draw (v1)--(plusv1);
\draw (betaN)--(plusvN); \draw (vN)--(plusvN);

\node[draw,fill=blue!20,minimum width=6cm,minimum height=1cm] (transformer) at (4,4) {transformer};
\draw (transformer.south -| plusu1.north)--(plusu1.north);
\draw (transformer.south -| plusuN.north)--(plusuN.north);
\draw (transformer.south -| plusv1.north)--(plusv1.north);
\draw (transformer.south -| plusvN.north)--(plusvN.north);

\node[draw,minimum height=0.6666666666666666cm,minimum width=0.6666666666666666cm] at (3.5,6.5) {$A^b$};
\node[draw,minimum height=0.6666666666666666cm,minimum width=0.6666666666666666cm,pattern=north east lines] at (4.5,6.5) {};
\node[draw,minimum height=0.6666666666666666cm,minimum width=0.6666666666666666cm,pattern=north east lines] at (3.5,7.5) {};
\node[draw,minimum height=0.6666666666666666cm,minimum width=0.6666666666666666cm] at (4.5,7.5) {$A^t$};
\node[draw=none,minimum height=1.3333333333cm,minimum width=1.3333333333cm] (am) at (4,7) {};

\draw (transformer)--(am);

\end{tikzpicture}
}
\caption{The transformer model we use for DeepEMD (see \S\ref{sec:deepEMD}) predicts directly the bipartite graph as an attention matrix.}\label{fig:transformer}
\end{minipage}
\end{figure}
We are interested in building a model which operates on a pair of point clouds $(U,V)$ as input, where $U, V \in \mathcal{D}^N$, $U = \{u_i\}_{i=1}^N$, $V = \{v_j\}_{j=1}^N$, $u_i, v_j \in \mathbb{R}^D$, and $N$ is the cardinality of the point clouds. Note that the points are unordered and the indexing is arbitrary. We denote the earth mover's distance between them as $d=\text{EMD}(U,V)$ where $d \in \mathbb{R}$. The goal of the model is to predict $d$ and $\nabla d$. Also, let $M \in \{0,1\}^{N \times N}$ denote the ground truth bipartite matching from EMD, where $M_{i,j}=1$ indicates that $u_i$ is matched to $v_j$ and vice-versa. Bipartite-ness implies   $\forall j, \sum_i M_{i,j}=1$ and $\forall i, \sum_j M_{i,j}=1$.

Since point clouds are unordered and invariant to elementwise permutation, we seek mappings $f: \mathcal{D}^N \times \mathcal{D}^N \rightarrow \mathbb{R}$ which are permutation invariant for any permutations $\pi$ and $\pi'$, i.e.,
\begin{align}
    f(U, V) = f(\pi(U), \pi'(V)),
\end{align}

In the following sections, we first introduce a simple MLP based baseline, followed by our transformer-based model, \emph{DeepEMD}.

\subsection{Predicting the distance}
\label{sec:baseline}
We propose a simple MLP baseline composed of a point-wise MLP backbone, followed by a prediction head which is also a MLP (see figure \ref{fig:mlp}). The backbone MLP takes a point cloud and returns an embedding $e \in \mathbb{R}^d$ as,
\begin{equation}
    e_u = \sum_{i=1}^n g(u_i), \quad e_v = \sum_{j=1}^n g(v_i)
\end{equation}
The prediction head then produces the final prediction as,
\begin{align}
    \hat{d} = h \left( e_u \oplus e_v \right) + h \left( e_v \oplus e_u \right),
    \label{eq:mlp_head}
\end{align}
where $\oplus$ denotes vector concatenation. Both $g$ and $h$ are composed of sequential linear layers with ReLU non-linearity between layers. The embeddings are permutation equivariant because of the sum aggregation which does not depend on the ordering of points. Further, we concatenate the embeddings both ways as in Eq. (\ref{eq:mlp_head}), which makes the mapping symmetric. We train the model with mean-squared error loss, $l = (d-\hat{d})^2$. Since the model does not predict the matching, we can interpret it from the direction of the gradient of a point $\delta v_j = \left[\frac{\partial \hat{d}}{\partial V}\right]_j$, e.g., by taking cosine similarity between $\delta v_j$ and $u_i - v_j$, where $u_i$ is the point matching to $v_j$ from EMD.


\subsection{Predicting bipartite matching}
\label{sec:deepEMD}

The transformer \cite{vaswani2017attention} intuitively seems to be a very good model for reasoning with point clouds and matching. Moreover, by considering that the output is the last layer's attention matrix, we can use it to directly predict the bipartite matching. Since the EMD--and consequently its gradient with respect to the point positions--is a function of the point positions and the matching array, predicting the latter leads to an accurate estimate of the former, that in particular is shielded from the issue of a possible decoupling between matching a functional point-wise (e.g. for MSE) and matching the gradients. While it is viable to directly predict the distance with a transformer-like architecture, we chose to only predict the matching since it is straightforward to estimate the distance using the predicted matching.

We propose \emph{DeepEMD} composed of a sequence of multi-head full attention layers, followed by a prediction head which is also a full attention layer, but with a single head. Given two point clouds $U$ and $V$, add a learned cloud-specific positional embedding to indicate if a point originates from $U$ or $V$,  we concatenate the points sequence and feed the resulting $I = U \cup V$ as input to the model. The group-id embedding helps the model in modulating attention locally within a point cloud as well as globally across both point clouds. We tried other variants with self-attention layers, cross attention layers, and an alternating mixture of both, but found full attention over both point clouds to be best performing.

For our problem, we get the input $X$ for the transformer by adding positional embeddings to $I$. Let $\vec{\mathit{0}}^n$ and $\vec{\mathit{1}}^n$ denote a vector of $n$ zeros and ones, respectively. $X$ is then obtained as,
\begin{equation}
    P = \vec{\mathit{0}}^n \cup \vec{\mathit{1}}^n, \quad
    X = I + W^P[P] \quad \text{(indexing)}
\end{equation}
The intermediate feature $t(X)$ from the transformer encoder (see appendix for details) has the same number of elements as $X$ with each element now being a contextualized representation for the corresponding point in the input. Further, these intermediate representation are fed into a single-head attention layer which outputs the attention matrix as,
\begin{gather}
    {K} = t(X){W}^K, \quad
    {Q} = t(X){W}^Q, \quad
    {A} = \frac{{Q}{K}^\top}{d_k}\\
    {A^t} = A_{:n, n:} \quad
    {A^b} = A_{n:, :n}
\end{gather}

Here, $A$ is a $2N \times 2N$ matrix and we slice the top-right block (first $N$ rows and last $N$ columns) as $A^t$ and bottom-left block (last $N$ rows and first $N$ columns) as $A^b$. $A^t_{i, j}$ can be interpreted as the relatedness of $u_i$ with $v_j$. Similarly, $A^b_{i, j}$ can be interpreted as the relatedness of $v_i$ with $u_j$.

Given $M$, the ground truth bipartite matching from EMD, we define the loss as the average of the cross-entropies as,
\begin{align}
    l(U, V) = \frac{1}{N} \sum_{i=1}^N\text{CE}({A}^t_{i, .}, M_{i, .})
     + \frac{1}{N}\sum_{i=1}^N\text{CE}({A}^b_{i, .}, M_{., i})
\end{align}

The EMD is then estimated with the predicted matching as,
\begin{gather}
    \phi^b(i) = \argmax_j {A}^b_{i,j}, \quad
    \phi^t(i) = \argmax_j {A}^t_{i,j} \\
    \hat{d}  = \frac{1}{2} \left( \sum_{i} \|u_i-v_{\phi^t(i)}\| + \sum_{i} \|v_i-u_{\phi^b(i)} \| \right)
\end{gather}


\section{Experiments}
\label{sec:experiment}

\begin{figure*}[t]
\centering
\begin{subfigure}[b]{.28\linewidth}
\includegraphics[width=\linewidth, clip]{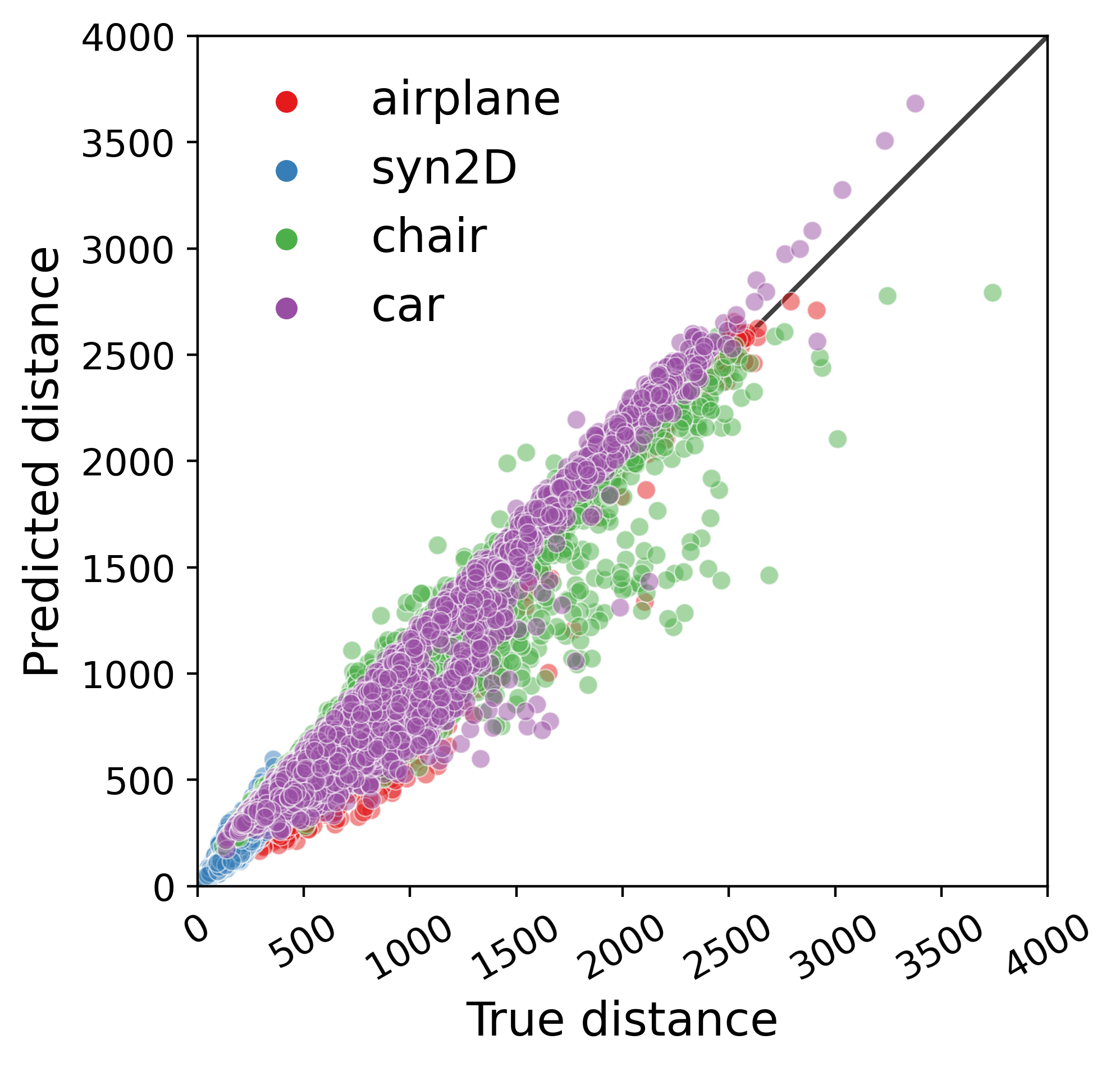}
\caption{Chamfer}
\label{fig:val_scatter_chamfer}
\end{subfigure}
\begin{subfigure}[b]{.28\linewidth}
\includegraphics[width=\linewidth]{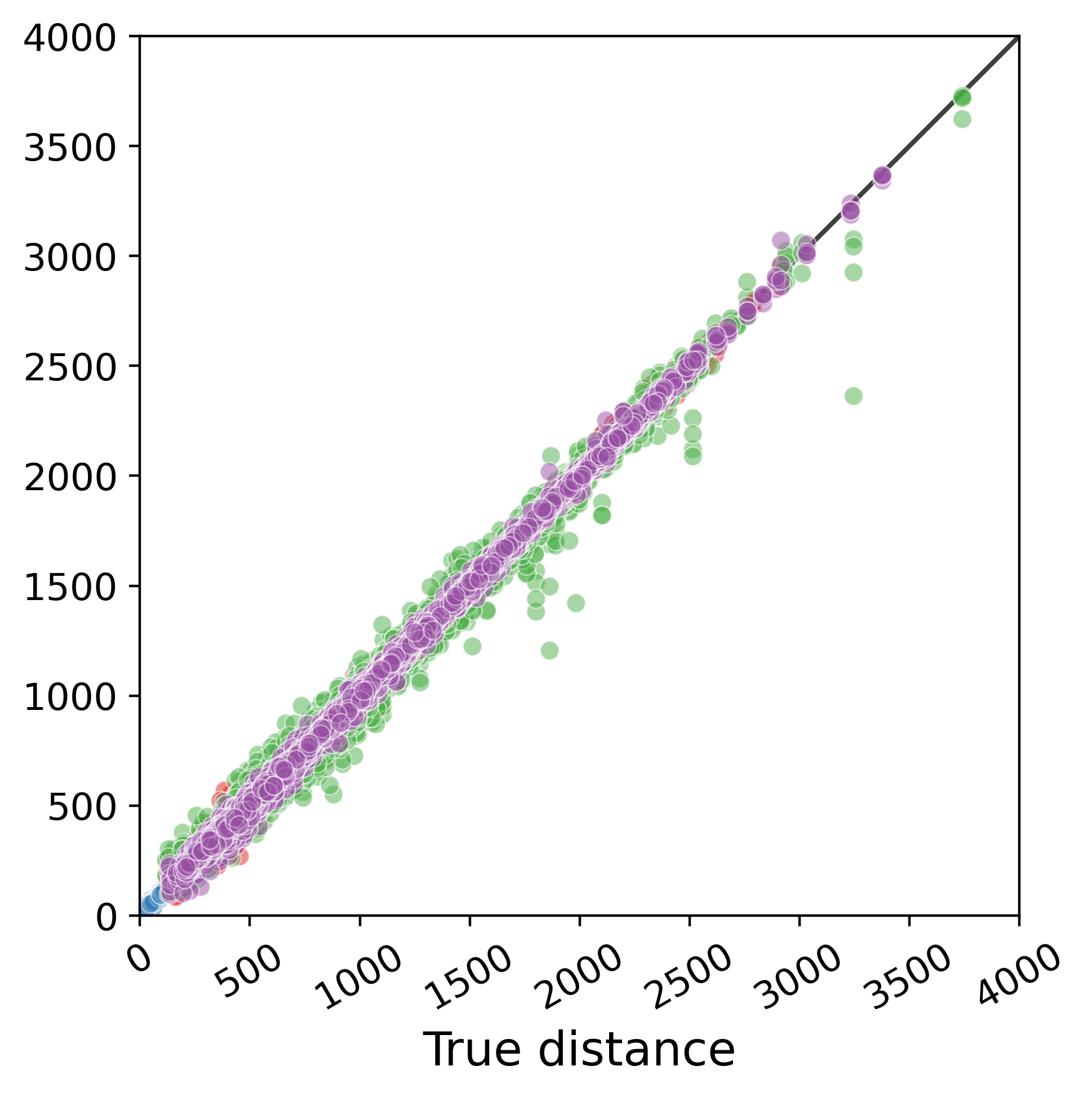}
\caption{MLP (ours)}\label{fig:val_scatter_mlp}
\end{subfigure}
\begin{subfigure}[b]{.28\linewidth}
\includegraphics[width=\linewidth]{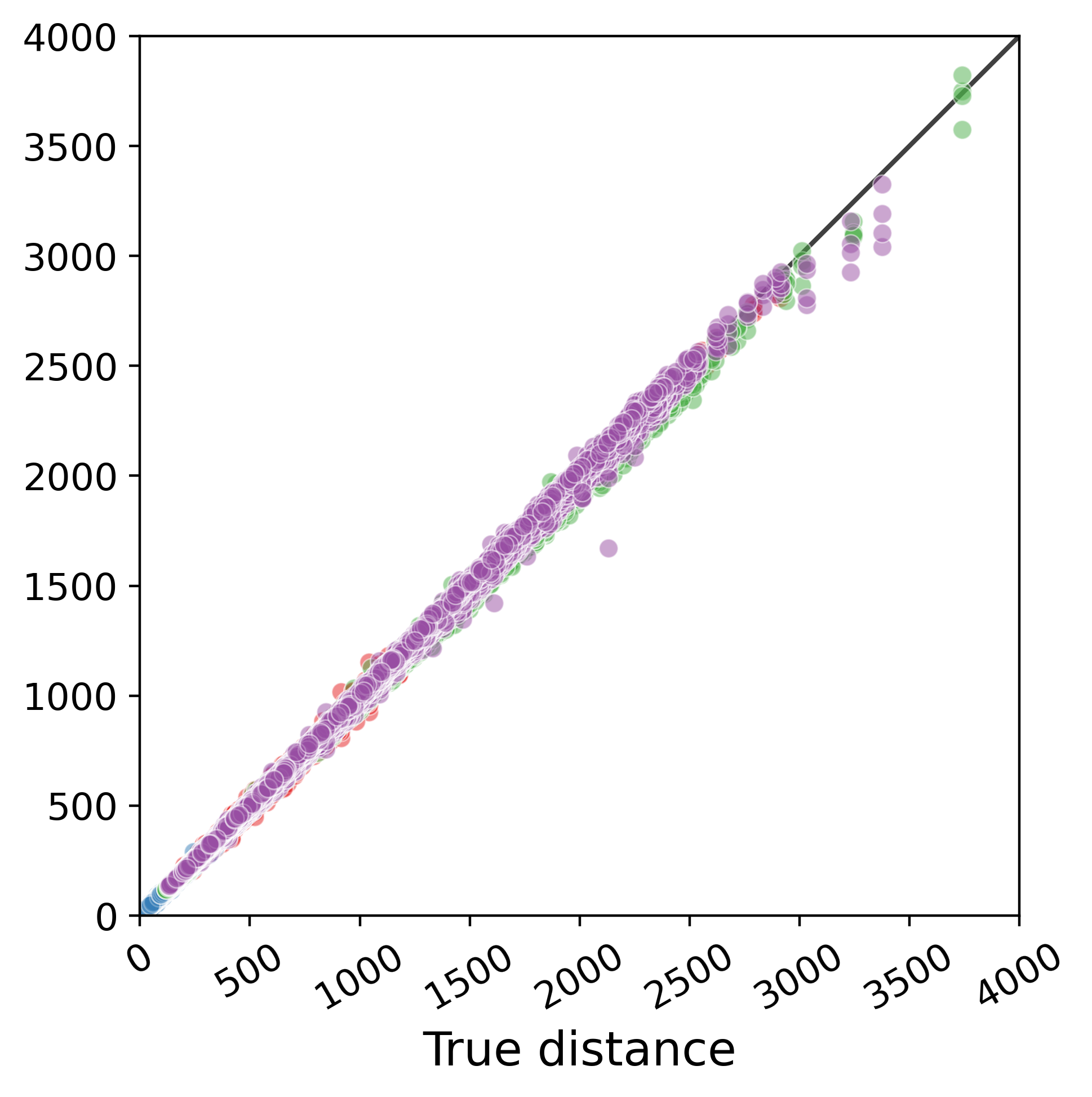}
\caption{DeepEMD (ours)}\label{fig:val_scatter_deepemd}
\end{subfigure}
\begin{subfigure}[b]{.14\linewidth}
\includegraphics[width=\linewidth]{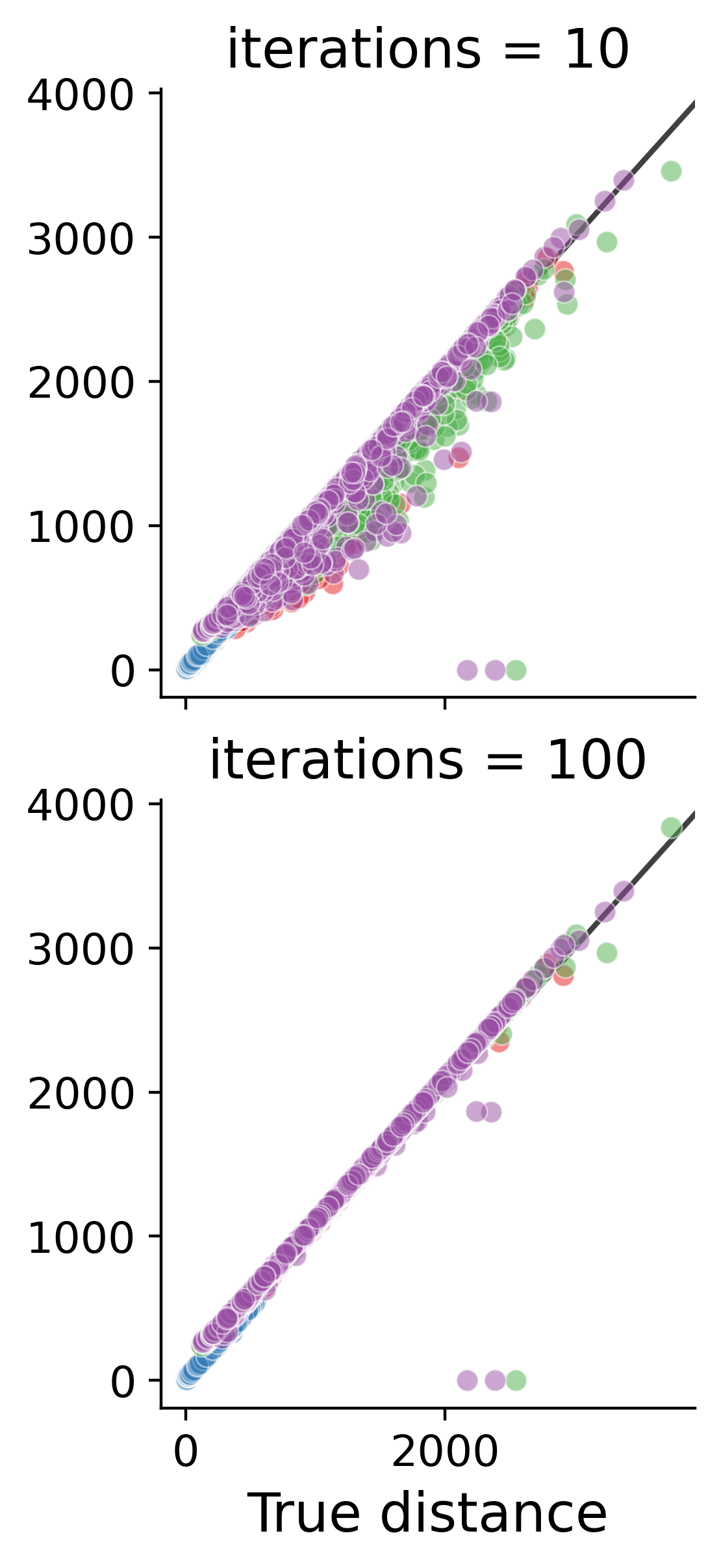}
\caption{Sinkhorn}\label{fig:val_scatter_sinkhorn}
\end{subfigure}
\caption{Scatter plot for true vs. approximate EMD from different models/metrics on validation splits for Syn2D and ShapeNet datasets. DeepEMD (ours) consistently performs better across different categories as it has less dispersion. Sinkhorn algorithm becomes more accurate with more iterations. Also note that it encounters numerical errors for some examples.}
\label{fig:val_scatter}
\end{figure*}

\begin{figure*}[t]
\centering
\begin{subfigure}[b]{.24\linewidth}
\includegraphics[width=1.05\linewidth]{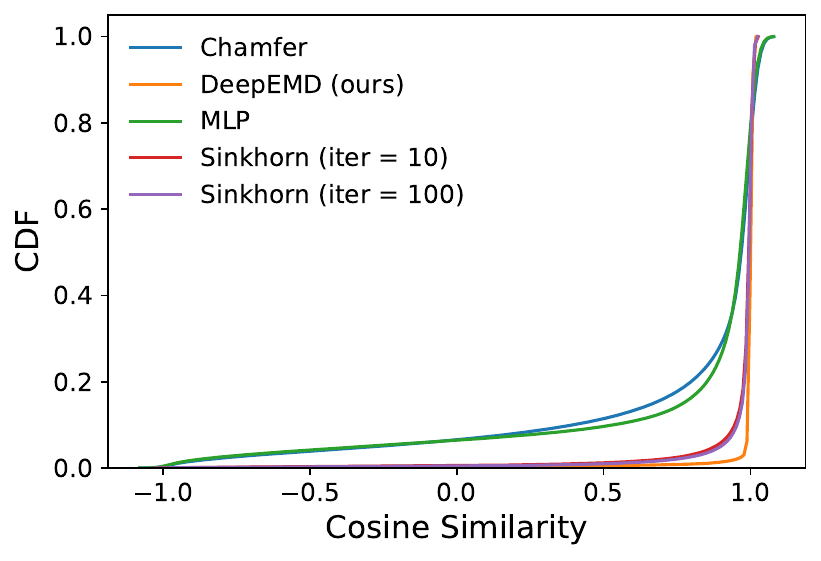}
\caption{Syn2D}\label{fig:val_grads_cos_sim_syn2D}
\end{subfigure} \:
\begin{subfigure}[b]{.24\linewidth}
\includegraphics[width=\linewidth]{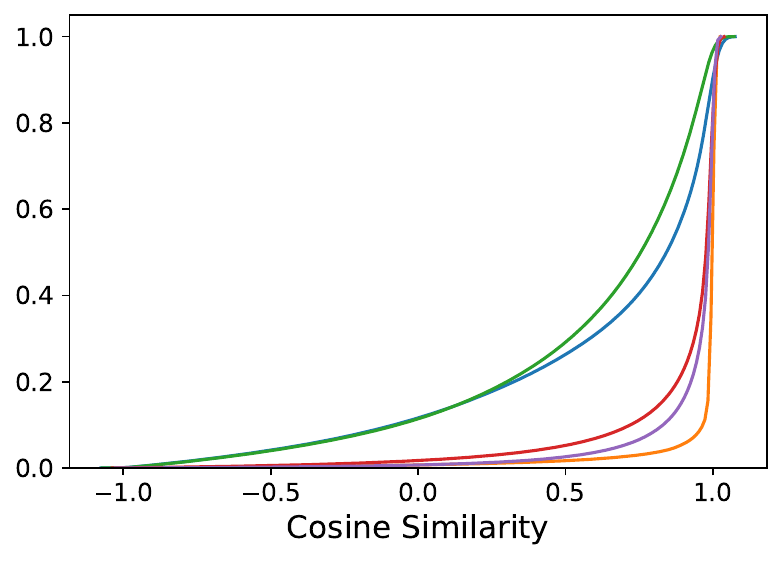}
\caption{ShapeNet Airplane}\label{fig:val_grads_cos_sim_airplane}
\end{subfigure}
\begin{subfigure}[b]{.24\linewidth}
\includegraphics[width=\linewidth]{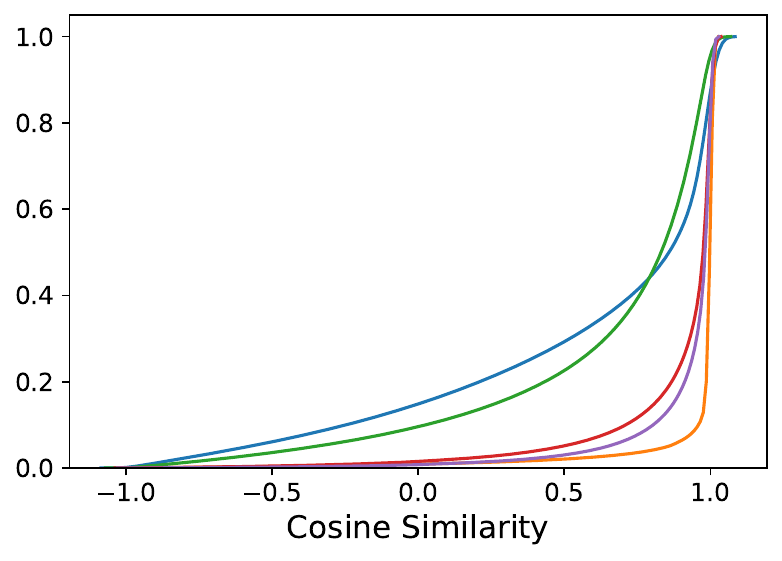}
\caption{ShapeNet Car}\label{fig:val_grads_cos_sim_car}
\end{subfigure}
\begin{subfigure}[b]{.24\linewidth}
\includegraphics[width=\linewidth]{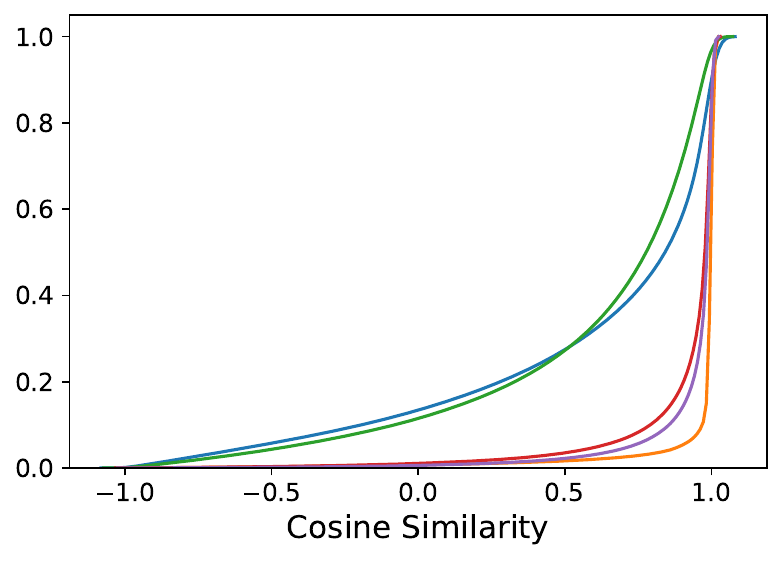}
\caption{ShapeNet Chair}\label{fig:val_grads_cos_sim_chair}
\end{subfigure}
\caption{CDF of cosine similarity between true and estimated gradients for all points across all point clouds collected together on validation splits for Syn2D and ShapeNet datasets. The ideal cdf curve should have all the mass at cosine similarity $1$. DeepEMD (ours) consistently outperforms all the other methods across different datasets.}
\label{fig:val_grads_cos_sim}
\end{figure*}

In this section, we present the overall experimental setup, performance results and comparisons of DeepEMD across various tasks.

\subsection{Datasets}
\label{sec:datasets}
We consider different datasets for our experiments - \emph{Syn2D}, ShapeNet \cite{chang2015shapenet}, ModelNet40 \cite{wu20153d} and ScanObjectNN \cite{uy2019revisiting}. \emph{Syn2D} consists of 2D point clouds generated synthetically by sampling points on squares and circles (see Fig \ref{fig:example_pcs}). ShapeNet and ModelNet40 are datasets of 3D point clouds derived from 3D CAD models for different real world objects like chairs, cars, airplanes, etc. ScanObjectNN is a relatively new real-world point cloud object dataset based on scanned indoor scene data. In order to improve and assess generalization, we augment the train and test splits  with synthetic perturbations. We provide more details about the datasets and these augmentations in the Appendix \S~\ref{sec:app_datasets}.

\subsection{Performance Measures}
\label{sec:measures}
We consider various measures to assess EMD approximation methods for distance as well as matching estimation. We compare accuracy and computation time to that of Sinkhorn and CD (see \S~\ref{sec:related}).

\paragraph{Distance Estimation.} We visualize the true vs. predicted distance through scatter plots (Fig. \ref{fig:val_scatter}), we expect the data points to be close to $x=y$ line. We compare various correlation measures : linear correlation $(r)$, Spearman correlation ($\rho$) and Kendall-Tau correlation $(\tau)$, to assess the quality of distance estimation. The Spearman and Kendall-Tau are rank-statistic based correlation measures, indicative of the correspondence between two rankings. Note that, correlation measures are useful metrics as they indicate appropriateness of the predicted metric as a distance measure, irrespective of their absolute values. Additionally we  look at different quantiles (RE$_n$) of relative approximation error, which penalizes the difference between absolute values of the predicted and true distance. 

\paragraph{Matching Estimation.} In order to assess quality of the matching, we consider the cosine similarity between the true and predicted gradient. The true gradient of EMD is always along the matched point. We visualize the cumulative distribution function (cdf) of cosine similarities (Fig. \ref{fig:val_grads_cos_sim}), where we expect all the mass to be close to $1$. We also look at different quantiles (CS$_n$) of the cosine similarity. We also consider accuracy which is computed as the average accuracy of matching source points to target points and vice-versa, bipartiteness (B) which is fraction of points with bipartite matching, and also bipartiteness-correctness (B$_{corr}$) which is fraction of points which are bipartite as well as matched correctly.

\subsection{Results}
\label{sec:results}

\paragraph{EMD Prediction.}

Fig. \ref{fig:val_scatter} shows the scatter plot of the true EMD vs. approximate EMD predicted from our trained models on the validation split for \emph{Syn2D} and ShapeNet datasets. Note that the validation split also contains the augmentations as discussed in Sec. \ref{sec:datasets}. We also validate on specific splits for which the results are shown in the appendix. The plots indicate that both DeepEMD (Fig. \ref{fig:val_scatter_deepemd}) and MLP baseline (Fig. \ref{fig:val_scatter_mlp}) approximate the EMD faithfully. The MLP baseline seems to struggle a bit on ShapeNet Chair dataset. The higher dispersion in Chamfer (Fig. \ref{fig:val_scatter_chamfer}) and Sinkhorn with $10$ iterations (Fig. \ref{fig:val_scatter_sinkhorn}, top) indicates poor EMD estimation. The approximation with Sinkhorn algorithm becomes more accurate with higher number of iterations (Fig. \ref{fig:val_scatter_sinkhorn}, bottom), as expected. 

We summarize various metrics in Tables \ref{tab:table_main_comparison}, \ref{tab:table_main_comparison_cate_dist} and \ref{tab:table_main_comparison_cate_matching} the appendix. DeepEMD and MLP baseline both achieve linear correlation higher than $0.99$ in each case. The models achieve Kendall-Tau correlation close to $0.99$ and $0.96$, respectively, and Spearman correlation close to $0.99$ in each case, indicating that ordering of the samples based on approximate distances and true distances are very similar, and monotonocity of samples are preserved. It can be observed that DeepEMD is best except for relative error where it may do worse than our method with MLP. Also it is interesting to note that Sinkhorn performs worse than DeepEMD on  correlation- or relative error-based measures.

\paragraph{Matching/Gradient Prediction.}

\begin{figure*}[t]
\centering
\begin{subfigure}[b]{.32\linewidth}
\includegraphics[width=1.04\linewidth]{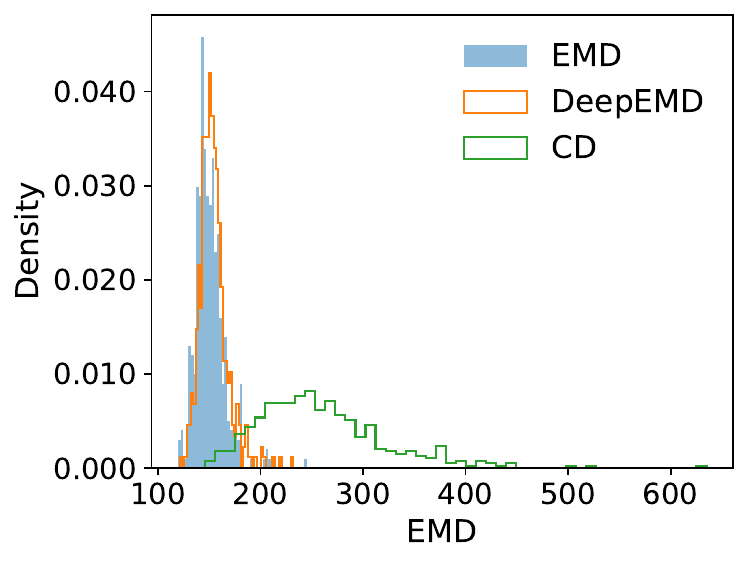}
\caption{ShapeNet Airplane}\label{fig:recon_emd_airplane}
\end{subfigure} \:
\begin{subfigure}[b]{.32\linewidth}
\includegraphics[width=1.0\linewidth]{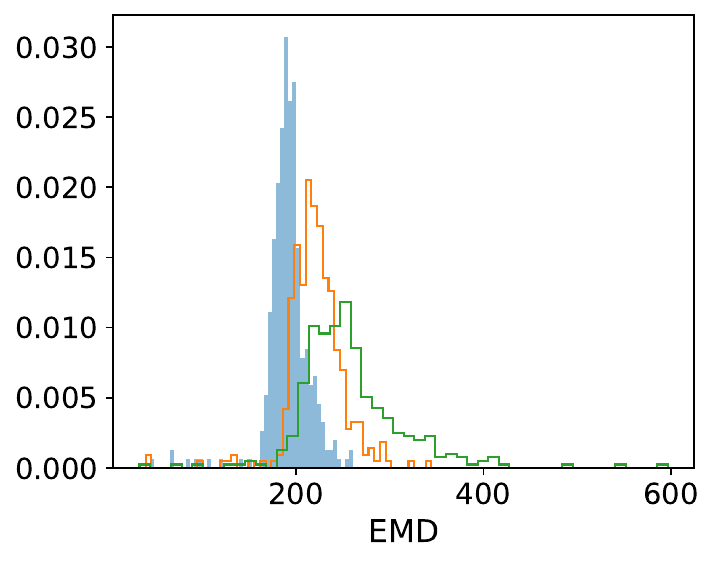}
\caption{ShapeNet Car}\label{fig:recon_emd_car}
\end{subfigure}
\begin{subfigure}[b]{.32\linewidth}
\includegraphics[width=1.0\linewidth]{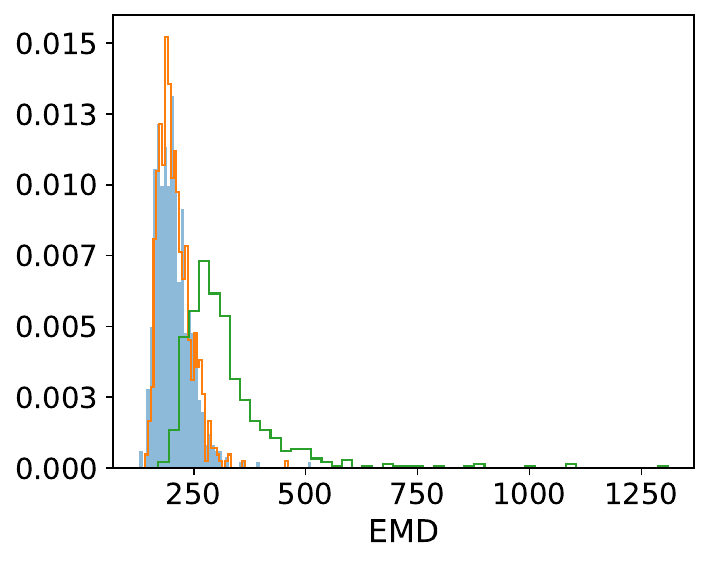}
\caption{ShapeNet Chair}\label{fig:recon_emd_chair}
\end{subfigure}
\caption{Comparison of EMD between input and reconstructed point clouds from SetVAE trained with different reconstruction losses. The better model should have smaller reconstruction loss and thus mass close to zero in the histograms. DeepEMD (ours) is consistently better as compared to Chamfer loss and very similar to EMD loss.}
\label{fig:recon_emd}
\end{figure*}

 Estimating the matching and gradient of the distance is particularly important for training models with DeepEMD as a surrogate distance function. Note that the gradient of a point from true EMD is always along the matched point in the other point cloud. Fig. \ref{fig:val_grads_cos_sim_syn2D} shows the cdf of cosine similarity between the true and estimated gradient for all the points across all point clouds collected together for \emph{Syn2D}, while Figs. \ref{fig:val_grads_cos_sim_airplane}, \ref{fig:val_grads_cos_sim_car}, and \ref{fig:val_grads_cos_sim_chair} for ShapeNet. The cdf  has most mass at cosine similarity close to $1$ with a very short tail and is never negative indicating that the estimated gradient is aligned with the true gradient for DeepEMD. This is particularly important when the model is used as a surrogate reconstruction loss. Ideally, the model should provide good estimate of the true gradient throughout training and more particularly in the very beginning when the reconstructions are very noisy, and also towards the end when reconstructions likely become very similar to the training distribution. We discuss more on this in the next sections and the appendix. The MLP baseline usually did not perform well in this regard and also in generalizing to unseen examples and thus was not useful for training generative models. The same issue can be observed with Chamfer distance as well. Table \ref{tab:table_main_comparison} (appendix) shows the cosine similarity quantiles (CS$_n$), as well as accuracy, bipartiteness (B) and bipartiteness-correctness (B$_{corr}$). DeepEMD performs better than other models and metrics in each of these measures and indicates better matching approximation.

\paragraph{Out-of-distribution generalization.}

The genralization of the prediction to a novel distribution is particularly importatnt for a surrogate metric. We test the out-of-distribution behaviour of our models in two different settings : Table \ref{tab:table_ood} shows the generalization performance of the model trained on a single category of ShapeNet and tested on validation split of multi-category ModelNet40 dataset, while 
Tables \ref{tab:table_ood_shapenet}, \ref{tab:ood-shapenet-dist} and \ref{tab:ood-shapenet-matching} in the appendix
show the performance when tested on different ShapeNet categories. The results indicate that DeepEMD generalizes well when test and train data differ without any adaptation or fine-tuning. Further, the validation performance on a category of a model trained on another category (see Appendix for details) is very similar to the performance of the model trained on the same category. These quite remarkable behaviors point towards the network “meta-learning” in some way the matching algorithm. This is further strengthened by the results on scaling to different number of points during test time as shown in Table \ref{tab:table_scaling}.
\begin{table}[!h]
\centering
\caption{Out-of-distribution (dataset) generalization for our models and comparison with other metrics (Chamfer and Sinkhorn), tested on full validation split for ModelNet40 (with 40 categories) and ScanObjectNN (with 15 categories). The models are trained on a single ShapeNet category. The reported numbers are averaged over these categories as well as four training seeds. The first six rows show distance estimation metrics (see \S~\ref{sec:measures}), while the last six rows correspond to matching estimation metrics. The arrows next to the metrics indicate whether higher ($\uparrow$) values are better or lower ($\downarrow$). Chamfer and Sinkhorn are deterministic, thus variances are not reported. Further, MLP does not provide accuracy and bipartiteness metrics.}
\label{tab:table_ood}
\begin{center}
\begin{sc}
\resizebox{1.0\columnwidth}{!}{%
\begin{tabular}{l|llll|llll}
\hline
dataset & \multicolumn{4}{c|}{ModelNet40}         & \multicolumn{4}{c}{ScanObjectNN} \\ 
\hline
model                   & chamfer     & sinkhorn        & mlp                       & deepemd                                        & chamfer           & sinkhorn            & mlp                   & deepemd               \\ \hline
$r$ ($\uparrow$)            &  $ 0.951 $    &  $ 0.971  $            &  $ 0.959 \pm 0.011 $      &  $ \mathbf{0.999} \pm 0.0 $      &  $ 0.971 $  &  $ 0.929$  &  $ 0.965 \pm 0.005 $  &  $ \mathbf{0.997} \pm 0.001 $  \\
$\rho$ ($\uparrow$)         &  $ 0.935 $    &  $ 0.988  $            &  $ 0.945 \pm 0.017 $      &  $ \mathbf{0.999} \pm 0.0 $      &  $ 0.979 $  &  $ 0.965$  &  $ 0.963 \pm 0.007 $  &  $ \mathbf{0.999} \pm 0.0 $    \\
$\tau$ ($\uparrow$)         &  $ 0.792 $    &  $ \mathbf{0.983}  $   &  $ 0.819 \pm 0.024 $      &  $ 0.974 \pm 0.002 $             &  $ 0.882 $  &  $ 0.968$  &  $ 0.855 \pm 0.011 $  &  $ \mathbf{0.973} \pm 0.002 $  \\
RE$_{0.1}$ ($\downarrow$)   &  $ 0.03 $     &  $ 0.057  $            &  $ 0.009 \pm 0.001 $      &  $ \mathbf{0.005} \pm 0.002 $    &  $ 0.025 $  &  $ 0.038$  &  $ 0.013 \pm 0.001 $  &  $ \mathbf{0.004} \pm 0.001 $  \\
RE$_{0.5}$ ($\downarrow$)   &  $ 0.129 $    &  $ 0.102  $            &  $ 0.062 \pm 0.005 $      &  $ \mathbf{0.019} \pm 0.004 $    &  $ 0.094 $  &  $ 0.078$  &  $ 0.076 \pm 0.005 $  &  $ \mathbf{0.019} \pm 0.004 $  \\
RE$_{0.9}$ ($\downarrow$)   &  $ 0.321 $    &  $ 0.2    $            &  $ 0.257 \pm 0.03 $       &  $ \mathbf{0.04} \pm 0.004 $     &  $ 0.282 $  &  $ 0.244$  &  $ 0.299 \pm 0.025 $  &  $ \mathbf{0.051} \pm 0.005 $  \\ \hline
CS$_{0.1}$ ($\uparrow$)     &  $ -0.067 $   &  $ 0.824  $            &  $ -0.293 \pm 0.047 $     &  $ \mathbf{0.927} \pm 0.003 $    &  $ 0.138 $  &  $ 0.879$  &  $ -0.208 \pm 0.042 $ &  $ \mathbf{0.946} \pm 0.002 $  \\
CS$_{0.5}$ ($\uparrow$)     &  $ 0.834 $    &  $ 0.986  $            &  $ 0.684 \pm 0.023 $      &  $ \mathbf{1.0} \pm 0.0 $        &  $ 0.917 $  &  $ 0.992$  &  $ 0.719 \pm 0.02 $   &  $ \mathbf{0.999} \pm 0.0 $    \\
CS$_{0.9}$ ($\uparrow$)     &  $ 0.997 $    &  $ 0.999  $            &  $ 0.96 \pm 0.003 $       &  $ \mathbf{1.0} \pm 0.0 $        &  $ 0.998 $  &  $ 1.0 $  &  $ 0.965 \pm 0.003 $  &  $ \mathbf{1.0} \pm 0.0 $      \\
Accuracy ($\uparrow$)       &  $ 12.651 $   &  $ 31.91  $            &  -                        &  $ \mathbf{56.38} \pm 0.604 $    &  $ 7.673 $  &  $ 20.04 $ &  -      &  $ \mathbf{40.671} \pm 0.62 $  \\
B ($\uparrow$)              &  $ 17.045 $   &  $ 33.458 $            &  -                        &  $ \mathbf{70.401} \pm 0.672 $   &  $ 9.474 $  &  $ 19.43 $ &  -      &  $ \mathbf{56.269} \pm 0.71 $  \\
B$_{corr}$ ($\uparrow$)     &  $ 6.544 $    &  $ 19.615 $            &  -                        &  $ \mathbf{47.084} \pm 0.741 $   &  $ 3.38  $   &  $ 9.961$ &  -      &  $ \mathbf{30.055} \pm 0.678 $ \\ \hline
\end{tabular}
}
\end{sc}
\end{center}
\end{table}
\paragraph{Scaling number of points.} Remarkably, the size of point clouds during testing can differ greatly from those during training without degrading performance. Table \ref{tab:table_scaling} shows performance of the model for test point cloud sizes ranging from $256$ to $8196$, while training was done with only $1024$ points. Prediction of the metric itself (top 6 rows) does not degrade for all practical purposes. Regarding the matching estimation, directional measure of performance related to the cosine similarity (rows CS$_{n}$) do not degrade neither. We can notice degradation in accuracy based measures (last 3 rows) which is natural since the problem becomes difficult with increasing number of points $N$ because of its combinatorial nature. For training when memory requirement is much higher due to backprop, we can use smaller number of points, and scale it up during inference without any fine-tuning.


\begin{table*}[!h]
\centering
\caption{Scaling number of points and out-of-distribution (scale) generalization for DeepEMD. The models are trained on a single ShapeNet category with 1024 points and tested on validation split of same category but with different number of points. Reported values are averaged over 4 training seeds. DeepEMD generalizes well to unseen number of points at test time without fine-tuning.}
\label{tab:table_scaling}
\begin{center}
\resizebox{1.0\textwidth}{!}{%
\begin{tabular}{l|lll|l|lll}
\hline
& \multicolumn{3}{c|}{$\longleftarrow$ Less \# points than training $\longrightarrow$}          &      \multicolumn{1}{|c|}{Trained}           & \multicolumn{3}{c}{$\longleftarrow$ More \# points than training $\longrightarrow$} \\
\hline
\# points & 256 & 512 & 768 & 1024 & 2048 & 4096 & 8192 \\ \hline
$r$ & $ 1.0 \pm 0.0$  & $ 1.0 \pm 0.0$  & $ 1.0 \pm 0.0$  & $ 1.0 \pm 0.0$  & $ 1.0 \pm 0.0$  & $ 0.999 \pm 0.0$  & $ 0.999 \pm 0.001$  \\
$\rho$ & $ 1.0 \pm 0.0$  & $ 1.0 \pm 0.0$  & $ 1.0 \pm 0.0$  & $ 1.0 \pm 0.0$  & $ 1.0 \pm 0.0$  & $ 0.999 \pm 0.0$  & $ 0.998 \pm 0.0$  \\
$\tau$ & $ 0.985 \pm 0.0$  & $ 0.987 \pm 0.0$  & $ 0.988 \pm 0.0$  & $ 0.988 \pm 0.001$  & $ 0.986 \pm 0.001$  & $ 0.981 \pm 0.002$  & $ 0.974 \pm 0.004$  \\
RE$_{0.1}$ & $ 0.002 \pm 0.001$  & $ 0.002 \pm 0.001$  & $ 0.004 \pm 0.002$  & $ 0.007 \pm 0.003$  & $ 0.012 \pm 0.005$  & $ 0.013 \pm 0.007$  & $ 0.014 \pm 0.008$  \\
RE$_{0.5}$ & $ 0.01 \pm 0.002$  & $ 0.011 \pm 0.002$  & $ 0.014 \pm 0.003$  & $ 0.017 \pm 0.005$  & $ 0.027 \pm 0.009$  & $ 0.034 \pm 0.013$  & $ 0.04 \pm 0.016$  \\
RE$_{0.9}$ & $ 0.026 \pm 0.003$  & $ 0.026 \pm 0.003$  & $ 0.029 \pm 0.004$  & $ 0.032 \pm 0.005$  & $ 0.042 \pm 0.009$  & $ 0.054 \pm 0.013$  & $ 0.066 \pm 0.018$  \\
\hline
CS$_{0.1}$ & $ 0.94 \pm 0.002$  & $ 0.955 \pm 0.002$  & $ 0.961 \pm 0.001$  & $ 0.964 \pm 0.001$  & $ 0.967 \pm 0.001$  & $ 0.967 \pm 0.001$  & -  \\
CS$_{0.5}$ & $ 1.0 \pm 0.0$  & $ 1.0 \pm 0.0$  & $ 1.0 \pm 0.0$  & $ 1.0 \pm 0.0$  & $ 1.0 \pm 0.0$  & $ 1.0 \pm 0.0$  & -  \\
CS$_{0.9}$ & $ 1.0 \pm 0.0$  & $ 1.0 \pm 0.0$  & $ 1.0 \pm 0.0$  & $ 1.0 \pm 0.0$  & $ 1.0 \pm 0.0$  & $ 1.0 \pm 0.0$  & -  \\
Accuracy & $ 72.348 \pm 0.44$  & $ 69.384 \pm 0.383$  & $ 66.901 \pm 0.379$  & $ 64.648 \pm 0.404$  & $ 57.588 \pm 0.464$  & $ 47.78 \pm 0.51$  & $ 35.274 \pm 0.483$  \\
B & $ 81.857 \pm 0.755$  & $ 80.101 \pm 0.547$  & $ 78.013 \pm 0.507$  & $ 75.896 \pm 0.521$  & $ 68.658 \pm 0.584$  & $ 58.109 \pm 0.597$  & $ 44.603 \pm 0.734$  \\
B$_{corr}$ & $ 64.838 \pm 0.756$  & $ 61.558 \pm 0.587$  & $ 58.545 \pm 0.547$  & $ 55.719 \pm 0.568$  & $ 46.831 \pm 0.618$  & $ 35.053 \pm 0.6$  & $ 21.606 \pm 0.469$  \\ \hline
\end{tabular}
}
\end{center}
\end{table*}
\paragraph{Computational Time and Complexity.}

\begin{figure*}[!ht]
\centering
\begin{subfigure}[b]{.49\linewidth}
\includegraphics[width=1.0\linewidth, trim=5 5 5 5, clip]{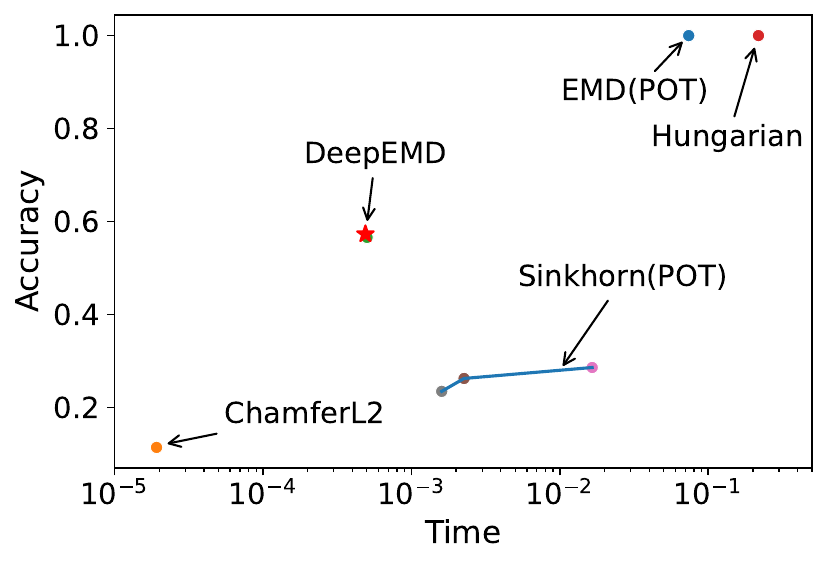}
\end{subfigure} \:
\begin{subfigure}[b]{.49\linewidth}
\includegraphics[width=1.0\linewidth, trim=5 5 5 5, clip]{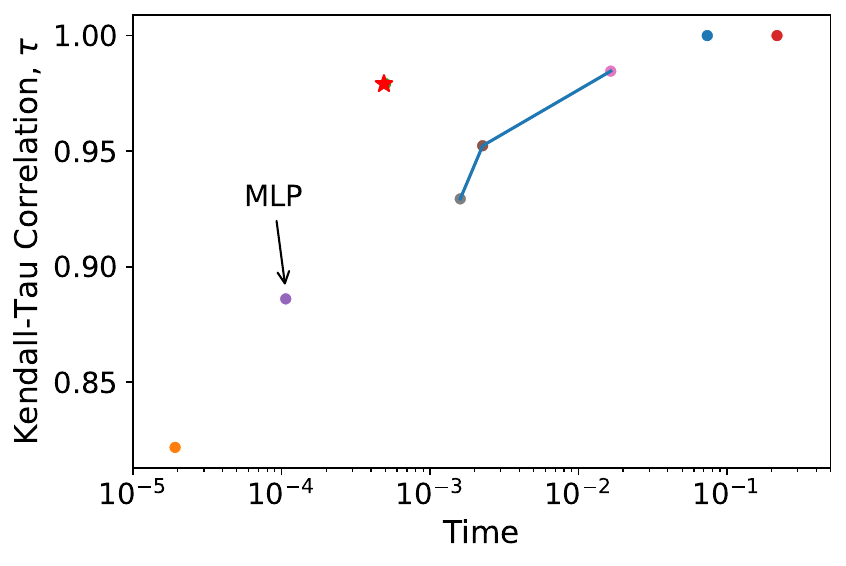}
\end{subfigure}
\caption{Comparison of empirical evaluation time and different performance measures : Accuracy (left) and Kendall-Tau correlation $\tau$ (right). We use Python Optimal Transport (POT) library for computing Sinkhorn distances, and show metrics at different iterations ($5$, $10$ and $100$). DeepEMD is $\sim100\times$  and $\sim40\times$ faster than Hungarian algorithm and Sinkhorn ($100$ iterations), respectively.}
\label{fig:eval_time}
\end{figure*}

Fig. \ref{fig:eval_time} compares the evaluation time for different models and metrics. DeepEMD achieves a significant speedup of about $100\times$ as compared to EMD and $40\times$ as compared to Sinkhorn with 100 iterations. This speedup becomes more pronounced on bigger point clouds as hungarian algorithm takes $O(N^3)$ time vs. $O(N^2)$ for DeepEMD.

\subsection{DeepEMD used as a loss}
\label{sec:genModels}
 Training a SetVAE, as for any auto-encoder, requires a reconstruction loss to assess the quality of the learned representation. While the eventual goal would be to minimize the EMD, standard approach uses Chamfer Distance due to the prohibitive computation cost of calculating the EMD. Instead of Chamfer Distance we propose to use DeepEMD and demonstrate its utility as a reconstruction loss as compared to Chamfer Distance.

DeepEMD was trained separately on each category of ShapeNet dataset and the trained model was then used as a surrogate reconstruction loss for training a variational auto-encoder. We use SetVAE \cite{kim2021setvae}, a transformer based VAE adapted for point clouds and set-structured data. The parameters of DeepEMD module are frozen during training of the SetVAE. We follow exactly the same protocol as in SetVAE and train using ShapeNet categories of airplane, chair, and car and also the same hyper-parameters for training. Fig. \ref{fig:val_recon_intro} and Fig. \ref{fig:val_recons_samples} (appendix) shows the reconstruction on validation data achieved by SetVAE models trained with different reconstruction losses. DeepEMD consistently achieves lower reconstruction EMD as compared to CD. This is further verified from Fig. \ref{fig:recon_emd} which shows the distribution of true EMD between a point cloud and its reconstruction.


\section{Conclusion and Future Work}
\label{sec:conclusion}
We propose DeepEMD, a method for fast approximation of EMD, improving time complexity from ${O}(N^3)$ to ${O}(N^2)$. It is composed of a multi-head multi-layer transformer, followed by a single-head full attention layer as the final output layer. It operates on two point clouds and outputs an attention matrix which is interpreted as the matching matrix and optimized to match the ground turh matching obtained from the hungarian algorithm. We demonstrated the effectiveness of DeepEMD in approximating the true EMD for synthetic 2D point clouds as well as real world datasets like ShapeNet, ModelNet40 and ScanObjectNN. It achieves a speed-up of $\times 100$ with $1024$ points. Further, we show that it estimates the gradients well, generalizes well for unseen point clouds (or distributions), and can be used for end-to-end training of point cloud autoencoders achieving faster convergence than Chamfer distance surrogate. 

It would be interesting to explore fast transformer variants to further improve from the quadratic time complexity for future work. In terms of architecture various pooling/un-pooling strategies can be explored which can help with both, better time complexity and improved feature learning. In this work, we estimate the $\text{Wasserstein}-2$ metric, and extension to other Wasserstein$-p$ metrics and other optimal transport problems could also be interesting for various applications. Lastly, extension to general probability distributions with fractional assignments (i.e. mass splitting) can also be very useful and valuable for some applications.

\bibliographystyle{plainnat}
\bibliography{main}

\begin{thebibliography}{21}
\providecommand{\natexlab}[1]{#1}
\providecommand{\url}[1]{\texttt{#1}}
\expandafter\ifx\csname urlstyle\endcsname\relax
  \providecommand{\doi}[1]{doi: #1}\else
  \providecommand{\doi}{doi: \begingroup \urlstyle{rm}\Url}\fi

\bibitem[Achlioptas et~al.(2018)Achlioptas, Diamanti, Mitliagkas, and Guibas]{achlioptas2018learning}
Panos Achlioptas, Olga Diamanti, Ioannis Mitliagkas, and Leonidas Guibas.
\newblock Learning representations and generative models for 3d point clouds.
\newblock In \emph{International conference on machine learning}, pages 40--49. PMLR, 2018.

\bibitem[Amos et~al.(2022)Amos, Cohen, Luise, and Redko]{amos2022meta}
Brandon Amos, Samuel Cohen, Giulia Luise, and Ievgen Redko.
\newblock Meta optimal transport.
\newblock \emph{arXiv preprint arXiv:2206.05262}, 2022.

\bibitem[Blondel et~al.(2018)Blondel, Seguy, and Rolet]{blondel2018smooth}
Mathieu Blondel, Vivien Seguy, and Antoine Rolet.
\newblock Smooth and sparse optimal transport.
\newblock In \emph{International conference on artificial intelligence and statistics}, pages 880--889. PMLR, 2018.

\bibitem[Chang et~al.(2015)Chang, Funkhouser, Guibas, Hanrahan, Huang, Li, Savarese, Savva, Song, Su, et~al.]{chang2015shapenet}
Angel~X Chang, Thomas Funkhouser, Leonidas Guibas, Pat Hanrahan, Qixing Huang, Zimo Li, Silvio Savarese, Manolis Savva, Shuran Song, Hao Su, et~al.
\newblock Shapenet: An information-rich 3d model repository.
\newblock \emph{arXiv preprint arXiv:1512.03012}, 2015.

\bibitem[Chuang et~al.(2022)Chuang, Jegelka, and Alvarez-Melis]{chuang2022infoot}
Ching-Yao Chuang, Stefanie Jegelka, and David Alvarez-Melis.
\newblock Infoot: Information maximizing optimal transport.
\newblock \emph{arXiv preprint arXiv:2210.03164}, 2022.

\bibitem[Cuturi(2013)]{cuturi2013sinkhorn}
Marco Cuturi.
\newblock Sinkhorn distances: Lightspeed computation of optimal transport.
\newblock \emph{Advances in neural information processing systems}, 26, 2013.

\bibitem[Flamary et~al.(2016)Flamary, Courty, Tuia, and Rakotomamonjy]{flamary2016optimal}
R~Flamary, N~Courty, D~Tuia, and A~Rakotomamonjy.
\newblock Optimal transport for domain adaptation.
\newblock \emph{IEEE Trans. Pattern Anal. Mach. Intell}, 1, 2016.

\bibitem[Flamary et~al.(2021)Flamary, Courty, Gramfort, Alaya, Boisbunon, Chambon, Chapel, Corenflos, Fatras, Fournier, et~al.]{flamary2021pot}
R{\'e}mi Flamary, Nicolas Courty, Alexandre Gramfort, Mokhtar~Z Alaya, Aur{\'e}lie Boisbunon, Stanislas Chambon, Laetitia Chapel, Adrien Corenflos, Kilian Fatras, Nemo Fournier, et~al.
\newblock Pot: Python optimal transport.
\newblock \emph{The Journal of Machine Learning Research}, 22\penalty0 (1):\penalty0 3571--3578, 2021.

\bibitem[Genevay et~al.(2018)Genevay, Peyr{\'e}, and Cuturi]{genevay2018learning}
Aude Genevay, Gabriel Peyr{\'e}, and Marco Cuturi.
\newblock Learning generative models with sinkhorn divergences.
\newblock In \emph{International Conference on Artificial Intelligence and Statistics}, pages 1608--1617. PMLR, 2018.

\bibitem[Huang et~al.(2022)Huang, Yang, Zhang, Cui, Zou, Chen, Zhao, and Liu]{huang2022learning}
Tianxin Huang, Xuemeng Yang, Jiangning Zhang, Jinhao Cui, Hao Zou, Jun Chen, Xiangrui Zhao, and Yong Liu.
\newblock Learning to train a point cloud reconstruction network without matching.
\newblock In \emph{Computer Vision--ECCV 2022: 17th European Conference, Tel Aviv, Israel, October 23--27, 2022, Proceedings, Part I}, pages 179--194. Springer, 2022.

\bibitem[Kim et~al.(2021)Kim, Yoo, Lee, and Hong]{kim2021setvae}
Jinwoo Kim, Jaehoon Yoo, Juho Lee, and Seunghoon Hong.
\newblock Setvae: Learning hierarchical composition for generative modeling of set-structured data.
\newblock In \emph{Proceedings of the IEEE/CVF Conference on Computer Vision and Pattern Recognition}, pages 15059--15068, 2021.

\bibitem[Kuhn(1955)]{kuhn1955hungarian}
Harold~W Kuhn.
\newblock The hungarian method for the assignment problem.
\newblock \emph{Naval research logistics quarterly}, 2\penalty0 (1-2):\penalty0 83--97, 1955.

\bibitem[Luo and Hu(2021)]{luo2021diffusion}
Shitong Luo and Wei Hu.
\newblock Diffusion probabilistic models for 3d point cloud generation.
\newblock In \emph{Proceedings of the IEEE/CVF Conference on Computer Vision and Pattern Recognition}, pages 2837--2845, 2021.

\bibitem[Shirdhonkar and Jacobs(2008)]{shirdhonkar2008approximate}
Sameer Shirdhonkar and David~W Jacobs.
\newblock Approximate earth mover’s distance in linear time.
\newblock In \emph{2008 IEEE Conference on Computer Vision and Pattern Recognition}, pages 1--8. IEEE, 2008.

\bibitem[Shu et~al.(2019)Shu, Park, and Kwon]{shu20193d}
Dong~Wook Shu, Sung~Woo Park, and Junseok Kwon.
\newblock 3d point cloud generative adversarial network based on tree structured graph convolutions.
\newblock In \emph{Proceedings of the IEEE/CVF international conference on computer vision}, pages 3859--3868, 2019.

\bibitem[Urbach et~al.(2020)Urbach, Ben-Shabat, and Lindenbaum]{urbach2020dpdist}
Dahlia Urbach, Yizhak Ben-Shabat, and Michael Lindenbaum.
\newblock Dpdist: Comparing point clouds using deep point cloud distance.
\newblock In \emph{European Conference on Computer Vision}, pages 545--560. Springer, 2020.

\bibitem[Uy et~al.(2019)Uy, Pham, Hua, Nguyen, and Yeung]{uy2019revisiting}
Mikaela~Angelina Uy, Quang-Hieu Pham, Binh-Son Hua, Thanh Nguyen, and Sai-Kit Yeung.
\newblock Revisiting point cloud classification: A new benchmark dataset and classification model on real-world data.
\newblock In \emph{Proceedings of the IEEE/CVF international conference on computer vision}, pages 1588--1597, 2019.

\bibitem[Vaswani et~al.(2017)Vaswani, Shazeer, Parmar, Uszkoreit, Jones, Gomez, Kaiser, and Polosukhin]{vaswani2017attention}
Ashish Vaswani, Noam Shazeer, Niki Parmar, Jakob Uszkoreit, Llion Jones, Aidan~N Gomez, {\L}ukasz Kaiser, and Illia Polosukhin.
\newblock Attention is all you need.
\newblock \emph{Advances in neural information processing systems}, 30, 2017.

\bibitem[Wu et~al.(2021)Wu, Pan, Zhang, Wang, Liu, and Lin]{wu2021density}
Tong Wu, Liang Pan, Junzhe Zhang, Tai Wang, Ziwei Liu, and Dahua Lin.
\newblock Density-aware chamfer distance as a comprehensive metric for point cloud completion.
\newblock \emph{arXiv preprint arXiv:2111.12702}, 2021.

\bibitem[Wu et~al.(2015)Wu, Song, Khosla, Yu, Zhang, Tang, and Xiao]{wu20153d}
Zhirong Wu, Shuran Song, Aditya Khosla, Fisher Yu, Linguang Zhang, Xiaoou Tang, and Jianxiong Xiao.
\newblock 3d shapenets: A deep representation for volumetric shapes.
\newblock In \emph{Proceedings of the IEEE conference on computer vision and pattern recognition}, pages 1912--1920, 2015.

\bibitem[Yang et~al.(2019)Yang, Huang, Hao, Liu, Belongie, and Hariharan]{yang2019pointflow}
Guandao Yang, Xun Huang, Zekun Hao, Ming-Yu Liu, Serge Belongie, and Bharath Hariharan.
\newblock Pointflow: 3d point cloud generation with continuous normalizing flows.
\newblock In \emph{Proceedings of the IEEE/CVF International Conference on Computer Vision}, pages 4541--4550, 2019.

\end{thebibliography}

\newpage

\appendix

\section{Optimal Transport and Wasserstein Distances}
\label{sec:app_ot}
The Wasserstein-$p$ metric between two probability distributions $\mu_X$ and $\nu_Y$ is defined as,
\begin{equation}
W_p(\mu, \nu)=\left(\inf _{\gamma \in \Gamma(\mu, \nu)} \mathbf{E}_{(x, y) \sim \gamma} \lVert x - y \rVert_p\right)^{1 / p},
\end{equation}
where $\Gamma(\mu, \nu)$ are all possible joint distributions where $X, Y \in \mathcal{D}$, $(\mathcal{D}, d)$ defines a metric space (here, $d=\lVert x - y \rVert_p$) and  marginals satisfy $\int_{\mathcal{D}} \gamma(x, y) \mathrm{d} y=\mu(x) $ and $\int_{\mathcal{D}} \gamma(x, y) \mathrm{d} x=\nu(y)$. Note that, while the distance is useful in itself, the optimal transport plan $\gamma*$ is also interesting for some applications. 

Given samples from $\mu$ and $\nu$, $W_p(\mu, \nu)$ can be computed by solving the optimal transport problem,
\begin{align}
\gamma^*= &\argmin_{\gamma \in \mathbb{R}_{+}^{m \times n}} \sum_{i, j} \gamma_{i, j} M_{i, j} \\
&\text { s.t. } \gamma 1=[\hat\mu]_m ; \gamma^T 1=[\hat\nu]_n ; \gamma \geq 0
\end{align}
where $[\hat\mu]_m$ and $[\hat\nu]_n$ represent binned histograms derived from samples from $\mu$ and $\nu$ with $m$ and $n$ bins respectively. $M$ is a $m \times n$ distance or cost matrix, $M_{i,j}$ represents the cost $d(x, y)$ to transport mass from bin $[\hat\mu]_m^i$ to bin $[\hat\nu]_n^j$.

It is also possible to avoid binning and compute the Wasserstein distance directly from samples. The problem can then be formulated as,
\begin{align}
\gamma^*=&\argmin_{\gamma \in \mathbb{R}_{+}^{n \times n}} \sum_{i, j} \gamma_{i, j} M_{i, j} \\
&\text { s.t. } \gamma 1=1 ; \gamma^T 1=1 ; \gamma \in \{0,1\}
\end{align}
where $M_{i,j}$ now represents the cost of transporting point $x_i$ to $y_j$. Each point is considered to be sampled i.i.d. from their respective distributions. Unlike the previous case, we can no longer transport a quanta of a point or mass i.e. fractional assignment is not meaningful, naturally leading to a bipartiteness constraint. This is exactly same as Eq (\ref{eq:emd}) and the optimal transport plan for the problem is a linear sum assignment problem and can be computed using the hungarian algorithm.

\section{Datasets}
\label{sec:app_datasets}
We train and evaluate on \emph{Syn2D}, ShapeNet and ModelNet40 datasets. 

\paragraph{\emph{Syn2D}.} We generate 2D synthetic point clouds by uniformly sampling $200$ points on simple 2D shapes of circles and squares. The circles are generated by uniformly sampling the center and radius from $(0,1]$. For the squares, we sample its center, rotation and scale uniformly from $[-0.5, 0.5]$, $[0, \frac{\pi}{2}]$, and $[0.5, 1]$, respectively. We refer to the synthetic dataset as \emph{Syn2D} (Fig. \ref{fig:example_pcs}).

\paragraph{ShapeNet.} ShapeNet \cite{chang2015shapenet} is a richly-annotated, large-scale dataset of 3D shapes. We train and evaluate our models using point clouds from one of the three categories in the ShapeNet dataset : airplane, chair, and car. We sample $1024$ points from the ShapeNet point clouds for training and evaluate on a range of point cloud sizes. 

\paragraph{ModelNet40.} \cite{chang2015shapenet} introduced the ModelNet project to provide a comprehensive and clean collection of $3D$ CAD models for objects, compiled using a list of the most common object categories. In our experiments, we evaluate our models trained with a single ShapeNet category on the ModelNet40 dataset. ModelNet40 has $40$ different categories.

\paragraph{ScanObjectNN.} \cite{uy2019revisiting} proposed the ScanObjectNN dataset, a real-world point cloud object dataset based on scanned indoor scene data, in order to provide a more realistic benchmark as compared to ModelNet40. In our experiments, we evaluate our models trained with a single ShapeNet category on pairs of point clouds derived from the training set of ScanObjectNN dataset.

Table \ref{tab:dataset_summary} provides a summary and statistics about of the datasets.
\begin{table}[!h]
\centering
\caption{Summary of datasets. The top five rows show statistics about the original point cloud datasets, while the bottom four rows show statistics about the pair point cloud dataset that was used for our evaluation and/or training. }
\label{tab:dataset_summary}
\begin{center}
\begin{sc}
\resizebox{1.\columnwidth}{!}{
\begin{tabular}{l|c|c|c|cccc}
\hline
                      & Syn2D & ModelNet40  & ScanObjectNN & ShapeNet &  Airplane     &  Car         & Chair       \\
\hline
\# Categories         & 2     & 40          &   15         & 55       & 1             & 1            & 1           \\
Feature dim           & 2     & 3           &   3        & 3        & 3             & 3            & 3           \\
Cardinality           & 200   & 2048        &   2048         & 15000    & 15000         & 15000        & 15000       \\
\# Train Samples      & 8000  & 9840        &   2309         & 35708    & 2832          & 2458         & 4612        \\
\# Val Samples        & 2000  & 2468        &   581         & 5158     & 405           & 352          & 662         \\    
\hline
\# Train Pair Samples & 20000 & -           &   -         & -        & 10000         & 10000        & 10000       \\
Train Cardinality     & 200   & -           &   -         & -        & 1024          & 1024         & 1024        \\
\# Val Pair Samples   & 5000  & 2000        &   10000         & -        & 2000          & 2000         & 2000        \\
Val Cardinality       & 200   & 1024        &   1024         & -        & 256-8192      & 256-8192     & 256-8192    \\
\hline
\end{tabular}
}
\end{sc}
\end{center}
\end{table}

Further, we build pairs $(U,V)$ of point clouds by randomly sampling pairs from the train or validation splits of point clouds datasets summarized in Table . We refer the first argument in the pair $U$ as the \emph{source} and the second argument $V$ as \emph{target}.

\subsection{Augmentations}
\label{sec:augmentation}
In order to improve generalization, we augment the datasets with point cloud pairs. We randomly sample a point cloud pair $(U, V)$ from the dataset and another noisy point cloud $N$ by randomly sampling points from $\mathcal{N}(0, 1)$ and scaling the whole point cloud by $\sigma \sim \mathcal{U}(0.1, 1.1)$.  Further, we augment the point cloud pairs according to the following schemes:
\begin{itemize}
    \item $(U, V)$ : the originally sampled pair.
    \item $(U, N)$ : target is replaced by the noisy point cloud.
    \item $(U, U + N)$ : target is a corrupted version of $U$ with additive noise $N$.
    \item $(U, \tilde U + N)$ : $\tilde U$ denotes a point cloud which is similar to $U$. For \emph{Syn2D}, we perturb the surface parameters (radius, scale, center, etc.) used for sampling $U$ and sample points on the perturbed surface. For ShapeNet, we independently sample different set of points from the original surface.
    \item $(U, V + N)$ : target is a corrupted version of a randomly sampled point cloud from the dataset.
 \end{itemize}

 The resulting dataset constitutes $20\%$ samples from each of the above splits. Validation splits are generated randomly and independently for \emph{Syn2D}, while for ShapeNet and ModelNet40, we use the validation split provided. We also augment the validation split with the same scheme as discussed above. The specifics of the pair point-cloud dataset is summarized in Table \ref{tab:dataset_summary}. 

\section{Models}

\paragraph{Transformers.}
Let $X = [x_1, \cdots, x_m] \in \mathbb{R}^{m \times d_{\text{model}}}$ be the input sequence (or set) of $m$ vectors. A transformer layer performs the following computation \cite{vaswani2017attention} :
\begin{gather*}
    {X}' = \text{LN}\left({X} + \text{Multihead}({X})\right)\\
    t_l({X}) =  \text{LN}\left({X}' + \text{FFN}({X}')\right)
\end{gather*}
where LN and FFN stand for layer norm and feed-forward network, respectively. Multihead denotes a multi-head attention layer which allows the model to jointly attend to information from different representation subspaces at different positions. It consists of a stack of $H$ scaled dot-product attention layers and computes key, query and value matrices, followed by a softmax as follows :
\begin{gather*}
    {K}_h = {X}{W}_h^K, \quad
    {Q}_h = {X}{W}_h^Q, \quad
    {V}_h = {X}{W}_h^V\\
    {A}_h = \text{soft-max}\left(\frac{{Q_h}{K_h}^\top}{d_k}\right){V_h}\\
    \text{MultiHead}({X}) = \Bar{{A}} = \text{concat}\left({A}_1, \cdots, {A}_H\right){W}_O
\end{gather*}

The final output of the encoder can be written as a composition:
\begin{gather*}
    t({X}) =  t_N\left(t_{N-1}\left(\ldots\left(t_1({X})\right)\right)\right)
\end{gather*}

\paragraph{MLP.} The baseline MLP model predicts directly the distance and it does not use the matching information. It has about $110K$ parameters in total. The point-wise MLP backbone $g(.)$ is composed of three hidden layer of sizes $4, 8$ and $16$, with ReLU non-linearity. It outputs a single embedding of dimension $128$ for each point cloud after aggregating the point level features. The embedding of the point clouds are then concatenated and passed to the prediction head which is also an MLP with four hidden layers  of sizes $256, 128, 64$ and $16$, and outputs a single scalar which is interpreted as the predicted distance. 

\paragraph{DeepEMD.}

We use a transformer encoder backbone which transforms the raw input point clouds into contextualized point level  features. The transformer encoder is followed by the output layer which computes the queries and keys and finally the attention matrix which is interpreted as the matching as explained in \S~\ref{sec:deepEMD}. The model constitutes about $803K$ learnable paramters, with $8$ transformer encoder layers, each with $6$ heads. The latent dimensions ($d_{\text{model}}, d_{\text{keys}}$, etc.) for each layer were all set to $78$.

We use the ADAM optimizer with a constant learning rate of $0.001$ for DeepEMD and $0.0001$ for the MLP.

\section{Sinkhorn Distance}
The Sinkhorn distance \cite{cuturi2013sinkhorn} considers a regularized OT optimization problem :
\begin{align*}
\gamma^*= &\argmin_{\gamma \in \mathbb{R}_{+}^{m \times n}} \sum_{i, j} \gamma_{i, j} M_{i, j} + \lambda\Omega(\gamma)\\
&\text { s.t. } \gamma 1=[\hat\mu]_m ; \gamma^T 1=[\hat\nu]_n ; \gamma \geq 0
\end{align*}
where $\lambda$ is the regularization coefficient and $O(\gamma) = \sum_{i, j} \gamma_{i,j} \log (\gamma_{i,j})$ is a entropy regularization term which makes the optimization problem smooth and strictly convex allowing for optimization procedures such as the Sinkhorn-Knopp algorithm. In this paper, we use the Python Optimal Transport (POT) python library \cite{flamary2021pot} for computing the Sinkhorn distances. It is an iterative algorithm and can be evaluated in $O(N^2)$ time complexity. 

In our experiments, we used a regularization multiplier of $0.1$. Reducing the multiplier typically resulted in numerical issues for a larger portion of the samples, whereas increasing it led to greater inaccuracies in estimation with the same computation time (for a fixed number of iterations). We show the performance metrics with different regularisation parameters in Table \ref{tab:sinkhorn_reg}.

\begin{table}[!ht]
    \centering
    \caption{Effect of regularization parameter on Sinkhorn's algorithm (100 iterations). Results on ScanObjectNN dataset with $10000$ samples. The second row shows number of instances which had numerical issues.
    }
    \label{tab:sinkhorn_reg}
    \begin{sc}
    \resizebox{0.4\columnwidth}{!}
    {%
    \begin{tabular}{l|llll}
\hline
$\lambda$      & 0.08 & 0.1   & 0.12 & 0.14                \\
\hline
\# Fail    & 686               & 80         & 6        & 0        \\
\hline
$r$        & $ 0.616 $  & $ 0.929  $     & $0.993 $   & $ 0.999 $    \\
$\rho$     & $ 0.708 $  & $ 0.965  $     & $0.997 $   & $ 0.999 $    \\
$\tau$     & $ 0.785 $  & $ 0.968  $     & $0.988 $   & $ 0.986 $    \\
RE$_{0.1}$ & $ 0.033 $  & $ 0.038  $     & $0.045 $   & $ 0.052 $    \\
RE$_{0.5}$ & $ 0.068 $  & $ 0.078  $     & $0.093 $   & $ 0.110 $    \\
RE$_{0.9}$ & $ 0.288 $  & $ 0.244  $     & $0.166 $   & $ 0.342 $    \\
CS$_{0.1}$ & $ 0.815 $  & $ 0.879  $     & $0.869 $   & $ 0.851 $    \\
CS$_{0.5}$ & $ 0.991 $  & $ 0.992  $     & $0.991 $   & $ 0.989 $    \\
CS$_{0.9}$ & $ 0.999 $  & $1.0$ & $0.999 $   & $ 0.999 $    \\
Accuracy   & $ 21.29 $  & $ 20.049  $    & $18.49 $   & $ 17.22 $    \\
B          & $ 20.72 $  & $ 19.439  $    & $18.09 $   & $ 17.03 $    \\
B$_{corr}$ & $ 10.97 $  & $ 9.961  $     & $8.98  $   & $ 8.23  $    \\
\hline
\end{tabular}
}
\end{sc}
\end{table}

The difficulty of choosing the regularisation parameter also makes it difficult for training generative models with Sinkhorn as it often leads to numerical issues. We experimented with several variants of Sinkhorn, including log-space Sinkhorn, log-stabilized Sinkhorn, Greenkhorn, etc. However, we discovered that they either exhibited significantly slower performance compared to the vanilla variant, frequently failed to converge to a satisfactory optimal transport matrix within a finite timeframe, or sometimes exhibited both issues.

\section{Extended Results}

\subsection{Distance and Matching Estimation}
We compare performance of different models and metrics in Table \ref{tab:table_main_comparison}. The models were trained on a single ShapeNet category and evaluated on the validation split of the same category. The numbers are averaged over all training categories as well as four training seeds. 
\begin{table}[!h]
\centering
\caption{Performance comparison of different metrics and models. The models are trained on a single ShapeNet category and evaluated on the test split of the same category. The reported numbers are averaged over all categories and four training seeds. The first six rows show distance estimation metrics (see \S~\ref{sec:measures}), while the last six rows correspond to matching estimation metrics. The arrows next to the metrics indicate whether higher ($\uparrow$) or lower ($\downarrow$) values are better. Chamfer and Sinkhorn are deterministic, thus variances are not reported. Further, our MLP model does not provide accuracy and bipartiteness metrics.}
\label{tab:table_main_comparison}
\begin{center}
\begin{sc}
\begin{tabular}{r|lllll}
\hline
model & chamfer & sinkhorn & mlp (ours) & deepemd (ours) \\ \hline
$r$ ($\uparrow$)            &  $ 0.963$  &  $ 0.995 $  &  $ 0.998 \pm 0.0 $  &  $ \mathbf{1.0} \pm 0.0 $  \\
$\rho$ ($\uparrow$)         &  $ 0.953 $  &  $ 0.997 $  &  $ 0.998 \pm 0.001 $  &  $ \mathbf{1.0} \pm 0.0 $  \\
$\tau$ ($\uparrow$)         &  $ 0.827  $  &  $ 0.987 $  &  $ 0.966 \pm 0.003 $  &  $ \mathbf{0.988} \pm 0.001 $  \\
RE$_{0.1}$ ($\downarrow$)   &  $ 0.023 $  &  $ 0.051 $  &  $ \mathbf{0.002} \pm 0.0 $  &  $ 0.007 \pm 0.003 $  \\
RE$_{0.5}$ ($\downarrow$)   &  $ 0.109 $  &  $ 0.106 $  &  $ \mathbf{0.015} \pm 0.001 $  &  $ 0.017 \pm 0.005 $  \\
RE$_{0.9}$ ($\downarrow$)   &  $ 0.31 $  &  $ 0.271 $  &  $ 0.076 \pm 0.006 $  &  $ \mathbf{0.032} \pm 0.005 $  \\ \hline
CS$_{0.1}$ ($\uparrow$)     &  $ -0.173  $  &  $ 0.831 $  &  $ -0.034 \pm 0.049 $  &  $ \mathbf{0.964} \pm 0.001 $  \\
CS$_{0.5}$ ($\uparrow$)     &  $ 0.85 $  &  $ 0.986 $  &  $ 0.798 \pm 0.018 $  &  $ \mathbf{1.0} \pm 0.0 $  \\
CS$_{0.9}$ ($\uparrow$)     &  $ 0.998 $  &  $ 0.999  $  &  $ 0.974 \pm 0.003 $  &  $ \mathbf{1.0} \pm 0.0 $  \\
Accuracy ($\uparrow$)       &  $ 11.677$  &  $ 28.407  $  &  -  &  $ \mathbf{64.648} \pm 0.404 $  \\
B ($\uparrow$)              &  $ 17.784$  &  $ 31.889 $  &  -  &  $ \mathbf{75.896} \pm 0.521 $  \\
B$_{corr}$ ($\uparrow$)     &  $ 5.626$  &  $ 16.658 $  &  -  &  $ \mathbf{55.719} \pm 0.568 $  \\
\hline
\end{tabular}
\end{sc}
\end{center}
\end{table}

Tables \ref{tab:table_main_comparison_cate_dist} and \ref{tab:table_main_comparison_cate_matching} show the  per-category performance comparison, averaged over four training seeds.
\begin{table*}[!h]
\centering
\caption{Per-category distance estimation performance measures of different models and metrics when train and test category are same. The reported number are averaged over four training seeds.}
\label{tab:table_main_comparison_cate_dist}
\begin{center}
\begin{sc}
\resizebox{1.05\textwidth}{!}{%
\begin{tabular}{ll|lll|lll}
\hline
\multicolumn{2}{l}{}                                         & $r$                            & $\rho$                         & $\tau$                          & RE$_{0.1}$                      & RE$_{0.5}$                      & RE$_{0.9}$                      \\ \hline
\multicolumn{1}{l|}{train\_cate}               & model/metric &                                &                                &                                 &                                 &                                 &                                 \\ \hline
\multicolumn{1}{l|}{\multirow{4}{*}{airplane}} & chamfer      & $0.9797$ & $0.9647$ & $0.8519$  & $0.0141$ & $0.0962$  & $0.3018$                     \\  
\multicolumn{1}{l|}{}                          & deepemd      & $ \mathbf{0.9998} \pm 0.0  $   & $ \mathbf{0.9997} \pm 0.0  $   & $ 0.9879 \pm 0.001  $  & $ 0.0039 \pm 0.0014  $          & $ 0.0142 \pm 0.0031  $          & $ \mathbf{0.0306} \pm 0.004  $  \\  
\multicolumn{1}{l|}{}                          & mlp          & $ 0.9992 \pm 0.0001  $         & $ 0.9986 \pm 0.0001  $         & $ 0.9722 \pm 0.0018  $          & $ \mathbf{0.002} \pm 0.0003  $  & $ \mathbf{0.0125} \pm 0.0018  $ & $ 0.0772 \pm 0.0038  $          \\ 
\multicolumn{1}{l|}{}                          & sinkhorn     & $0.9998$                       & $0.9997$                       & $\mathbf{0.9881}$                        & $0.0496$                        & $0.1104$                        & $0.2984$                       \\ \hline
\multicolumn{1}{l|}{\multirow{4}{*}{car}}      & chamfer      & $0.9675$                       & $0.9564$                       & $ 0.8318 $                      & $ 0.035 $                       & $0.1167$                         & $0.324$                          \\ 
\multicolumn{1}{l|}{}                          & deepemd      & $ \mathbf{0.9997} \pm 0.0001 $ & $ \mathbf{0.9998} \pm 0.0$     & $ \mathbf{0.9891} \pm 0.0006  $ & $ 0.006 \pm 0.0047  $           & $ 0.0156 \pm 0.0068  $          & $ \mathbf{0.0302} \pm 0.0097  $ \\ 
\multicolumn{1}{l|}{}                          & mlp          & $ 0.9993 \pm 0.0002  $         & $ 0.9988 \pm 0.0004  $         & $ 0.9746 \pm 0.0035  $          & $ \mathbf{0.0018} \pm 0.0002  $ & $ \mathbf{0.0112} \pm 0.001  $  & $ 0.0625 \pm 0.012  $           \\ 
\multicolumn{1}{l|}{}                          & sinkhorn     & $ 0.991 $  & $ 0.9941$  & $0.9854$  & $0.0531$  & $0.1124$  & $0.2986$                     \\ \hline
\multicolumn{1}{l|}{\multirow{4}{*}{chair}}    & chamfer      & $ 0.9431 $                     & $ 0.9382 $                     & $ 0.7976 $                      & $ 0.0192 $                      & $ 0.1133 $                      & $ 0.3037 $                     \\  
\multicolumn{1}{l|}{}                          & deepemd      & $ \mathbf{0.9997} \pm 0.0  $   & $ \mathbf{0.9997} \pm 0.0001 $ & $0.9866 \pm 0.0013  $ & $ 0.0103 \pm 0.0088  $          & $ 0.0225 \pm 0.0118  $          & $ \mathbf{0.0356} \pm 0.012  $  \\  
\multicolumn{1}{l|}{}                          & mlp          & $ 0.9965 \pm 0.0013  $         & $ 0.9959 \pm 0.0015  $         & $ 0.9504 \pm 0.0079  $          & $ \mathbf{0.0037} \pm 0.0006  $ & $ \mathbf{0.0214} \pm 0.0023  $ & $ 0.0881 \pm 0.0124  $          \\ 
\multicolumn{1}{l|}{}                          & sinkhorn     & $ 0.9942 $                     & $0.9969$                       & $\mathbf{0.9886}$                        & $0.049$                         & $0.0955$                        & $0.2167$                        \\ \hline
\end{tabular}%
}
\end{sc}
\end{center}
\end{table*}

\begin{table*}[]
\centering
\caption{Per-category matching estimation performance measures of different models and metrics when train and test category are same. The reported number are averaged over four training seeds.}
\label{tab:table_main_comparison_cate_matching}
\begin{center}
\begin{small}
\begin{sc}
\resizebox{1.05\textwidth}{!}{%
\begin{tabular}{|ll|lll|lll}
\hline
\multicolumn{2}{l}{}   & CS$_{0.1}$                      & CS$_{0.5}$                & CS$_{0.9}$                & accuracy                         & B                                & B\_corr                          \\ \hline
\multicolumn{1}{l|}{train\_cate}               & model/metric &                                &                                &                                 &                                 &                                 &                                 \\ \hline
\multicolumn{1}{l|}{\multirow{4}{*}{airplane}} & chamfer  & $-0.0813$  & $0.8446$ & $0.9973$  & $10.1768$  & $16.7437$  & $4.7461$                     \\  
         \multicolumn{1}{l|}{}                 & deepemd  & $ \mathbf{0.9643} \pm 0.0027  $ & $ \mathbf{1.0 \pm 0.0}  $ & $ \mathbf{1.0} \pm 0.0  $ & $ \mathbf{61.9119} \pm 0.9043  $ & $ \mathbf{73.3128} \pm 1.1746  $ & $ \mathbf{52.2082} \pm 1.246  $  \\ 
         \multicolumn{1}{l|}{}                 & mlp      & $-0.0492 \pm 0.0533$                               & $0.7766 \pm 0.0263$                         & $0.9722 \pm 0.003$                         & -                                & -                                & -                                \\  
         \multicolumn{1}{l|}{}                 & sinkhorn & $0.8314$  & $0.9871$  & $0.9994$ & $25.2956$  & $29.018$  & $13.9732$                      \\ \hline
\multicolumn{1}{l|}{\multirow{4}{*}{car}}      & chamfer  & $-0.246$  & $0.8615$  & $1.0$  & $13.7079$  & $20.7186$  & $6.8617$                       \\  
         \multicolumn{1}{l|}{}                 & deepemd  & $ \mathbf{0.9585} \pm 0.0025  $ & $ \mathbf{1.0 \pm 0.0}  $ & $ \mathbf{1.0} \pm 0.0  $ & $ \mathbf{67.7243} \pm 0.6554  $ & $ \mathbf{78.7286} \pm 0.797  $  & $ \mathbf{59.6723} \pm 0.9245  $ \\  
         \multicolumn{1}{l|}{}                 & mlp      & $0.017 \pm 0.0652$	& $0.8388 \pm 0.0187$	 & $0.9804 \pm 0.0029$       & -                                & -                                & -                                \\  
         \multicolumn{1}{l|}{}                 & sinkhorn & $0.8043$ & $0.9845$  & $0.9993$ & $31.1621$  & $35.5971$  & $19.3105$                    \\ \hline
\multicolumn{1}{l|}{\multirow{4}{*}{chair}}    & chamfer  & $-0.1929$  & $0.8442$ & $0.9976$  & $11.145$  & $15.8883$  & $5.2694$                      \\  
         \multicolumn{1}{l|}{}                 & deepemd  & $ \mathbf{0.9703} \pm 0.0007  $ & $ \mathbf{1.0 \pm 0.0}  $ & $ \mathbf{1.0} \pm 0.0  $ & $ \mathbf{64.3079} \pm 0.4712  $ & $ \mathbf{75.6459} \pm 0.657  $  & $ \mathbf{55.2757} \pm 0.7011  $ \\  
         \multicolumn{1}{l|}{}                 & mlp      & $-0.0695 \pm 0.1211$ &	$0.7793 \pm 0.0438$  &	$0.97 \pm 0.0072$             & -                                & -                                & -                                \\  
          \multicolumn{1}{l|}{}                & sinkhorn & $0.8558$  & $0.9876$  & $0.9992$  & $28.7644$  & $31.0521$  & $16.69$                       \\ \hline
\end{tabular}%
}
\end{sc}
\end{small}
\end{center}
\end{table*}

\subsection{Out-of-distribution generalization}

Table \ref{tab:table_ood_shapenet} shows the out-of-distribution generalization performance for our models. The trained model on a particular ShapeNet category is evaluated on the validation split of other ShapeNet categories. The numbers are averaged over these other categories as well as four training seeds. Tables \ref{tab:ood-shapenet-dist} and \ref{tab:ood-shapenet-matching} show the performance in the same setting but for each test category separately.

\begin{table}[!h]
\centering
\caption{Out-of-distribution (category) generalization for our models and comparison with other metrics (Chamfer and Sinkhorn). The models are trained on a single ShapeNet category and evaluated on other ShapeNet categories. The reported numbers are averaged over these categories as well as four training seeds. The first five rows show distance estimation metrics (see \S~\ref{sec:measures}), while the last five rows correspond to matching estimation metrics. The arrows next to the metrics indicate whether higher ($\uparrow$) or lower ($\downarrow$) values are better. }
\label{tab:table_ood_shapenet}
\begin{center}
\begin{sc}
\resizebox{0.9\columnwidth}{!}{%
\begin{tabular}{l|ll|ll}
\hline
model & mlp (ood) & mlp & deepemd (ood) & deepemd \\ \hline
$r$ ($\uparrow$)            & $0.98 \pm 0.019$ & $0.998 \pm 0.001$ & $0.999 \pm 0.001$ & $1.0 \pm 0.0$ \\
$\rho$ ($\uparrow$)         & $0.976 \pm 0.02$ & $0.998 \pm 0.001$ & $0.999 \pm 0.0$ & $1.0 \pm 0.0$ \\
$\tau$ ($\uparrow$)         & $0.886 \pm 0.04$ & $0.966 \pm 0.004$ & $0.977 \pm 0.003$ & $0.988 \pm 0.001$ \\
RE$_{0.1}$ ($\downarrow$)   & $0.014 \pm 0.003$ & $0.002 \pm 0.0$ & $0.009 \pm 0.009$ & $0.007 \pm 0.005$ \\
RE$_{0.5}$ ($\downarrow$)   & $0.065 \pm 0.018$ & $0.015 \pm 0.002$ & $0.024 \pm 0.012$ & $0.017 \pm 0.007$ \\
RE$_{0.9}$ ($\downarrow$)   & $0.319 \pm 0.132$ & $0.076 \pm 0.009$ & $0.05 \pm 0.011$ & $0.032 \pm 0.008$ \\ \hline
CS$_{0.1}$ ($\uparrow$)     & $-0.208 \pm 0.089$ & $-0.034 \pm 0.074$ & $0.933 \pm 0.004$ & $0.964 \pm 0.002$ \\
CS$_{0.5}$ ($\uparrow$)     & $0.714 \pm 0.047$ & $0.798 \pm 0.027$ & $1.0 \pm 0.0$ & $1.0 \pm 0.0$ \\
CS$_{0.9}$ ($\uparrow$)     & $0.963 \pm 0.006$ & $0.974 \pm 0.004$ & $1.0 \pm 0.0$ & $1.0 \pm 0.0$ \\
Accuracy ($\uparrow$)       & - & - & $54.35 \pm 1.16$ & $64.648 \pm 0.606$ \\
B ($\uparrow$)              & - & - & $67.922 \pm 1.343$ & $75.896 \pm 0.782$ \\
B$_{corr}$ ($\uparrow$)     & - & - & $44.293 \pm 1.445$ & $55.719 \pm 0.851$ \\ \hline
\end{tabular}
}
\end{sc}
\end{center}
\end{table} 

\begin{table*}[h]
\centering
\caption{Per-category distance estimation performance measures of different models and metrics when train and test category are different. The reported number are averaged over four training seeds.}
\label{tab:ood-shapenet-dist}
\begin{center}
\begin{small}
\begin{sc}
\resizebox{1.05\textwidth}{!}{%
\begin{tabular}{|ll|lll|lll}
\hline
\multicolumn{2}{l}{}                                       & $r$                   & $\rho$                & $\tau$                 & RE$_{0.1}$            & RE$_{0.5}$            & RE$_{0.9}$            \\ \hline
\multicolumn{1}{l|}{train\_cate}               & test\_cate &                       &                       &                       &                       &                       &                       \\ \hline
\multicolumn{1}{l|}{\multirow{3}{*}{airplane}} & airplane   & $ 0.9998 \pm 0.0 $    & $ 0.9997 \pm 0.0 $    & $ 0.9879 \pm 0.001 $  & $ 0.0039 \pm 0.0014 $ & $ 0.0142 \pm 0.0031 $ & $ 0.0306 \pm 0.004 $  \\ 
\multicolumn{1}{l|}{}                          & car        & $ 0.9989 \pm 0.0006 $ & $ 0.9997 \pm 0.0001 $ & $ 0.9864 \pm 0.001 $  & $ 0.0035 \pm 0.0005 $ & $ 0.016 \pm 0.0036 $  & $ 0.0308 \pm 0.0039 $ \\ 
\multicolumn{1}{l|}{}                          & chair      & $ 0.9981 \pm 0.0003 $ & $ 0.9983 \pm 0.0003 $ & $ 0.9689 \pm 0.0016 $ & $ 0.0044 \pm 0.0011 $ & $ 0.0212 \pm 0.0033 $ & $ 0.0491 \pm 0.0048 $ \\ \hline
\multicolumn{1}{l|}{\multirow{3}{*}{car}}      & airplane   & $ 0.9993 \pm 0.0002 $ & $ 0.9989 \pm 0.0003 $ & $ 0.9766 \pm 0.0017 $ & $ 0.0092 \pm 0.0065 $ & $ 0.0256 \pm 0.011 $  & $ 0.0634 \pm 0.0114 $ \\ 
\multicolumn{1}{l|}{}                          & car        & $ 0.9997 \pm 0.0001 $ & $ 0.9998 \pm 0.0 $    & $ 0.9891 \pm 0.0006 $ & $ 0.006 \pm 0.0047 $  & $ 0.0156 \pm 0.0068 $ & $ 0.0302 \pm 0.0097 $ \\ 
\multicolumn{1}{l|}{}                          & chair      & $ 0.9978 \pm 0.0001 $ & $ 0.9981 \pm 0.0002 $ & $ 0.9671 \pm 0.0017 $ & $ 0.0045 \pm 0.0019 $ & $ 0.0206 \pm 0.006 $  & $ 0.0525 \pm 0.0022 $ \\ \hline
\multicolumn{1}{l|}{\multirow{3}{*}{chair}}    & airplane   & $ 0.999 \pm 0.0004 $  & $ 0.9983 \pm 0.0008 $ & $ 0.9751 \pm 0.0047 $ & $ 0.0164 \pm 0.0114 $ & $ 0.034 \pm 0.0127 $  & $ 0.0631 \pm 0.0092 $ \\ 
\multicolumn{1}{l|}{}                          & car        & $ 0.9987 \pm 0.0007 $ & $ 0.9996 \pm 0.0001 $ & $ 0.9867 \pm 0.0007 $ & $ 0.0141 \pm 0.0116 $ & $ 0.0275 \pm 0.014 $  & $ 0.0422 \pm 0.0139 $ \\ 
\multicolumn{1}{l|}{}                          & chair      & $ 0.9997 \pm 0.0 $    & $ 0.9997 \pm 0.0001 $ & $ 0.9866 \pm 0.0013 $ & $ 0.0103 \pm 0.0088 $ & $ 0.0225 \pm 0.0118 $ & $ 0.0356 \pm 0.012 $  \\ \hline
\end{tabular}%
}
\end{sc}
\end{small}
\end{center}
\end{table*}

\begin{table*}[h]
\centering
\caption{Per-category matching estimation performance measures of different models and metrics when train and test category are different. The reported number are averaged over four training seeds.}
\label{tab:ood-shapenet-matching}
\begin{center}
\begin{small}
\begin{sc}
\resizebox{1.05\columnwidth}{!}{%
\begin{tabular}{ll|lll|lll}
\hline
\multicolumn{2}{l}{}                                       & CS$_{0.1}$            & CS$_{0.5}$            & CS$_{0.9}$      & accuracy               & B                      & B\_corr                \\ \hline
\multicolumn{1}{l|}{train\_cate}               & test\_cate &                       &                       &                 &                        &                        &                        \\ \hline
\multicolumn{1}{l|}{\multirow{3}{*}{airplane}} & airplane   & $ 0.9643 \pm 0.0027 $ & $ 1.0 \pm 0.0 $       & $ 1.0 \pm 0.0 $ & $ 61.9119 \pm 0.9043 $ & $ 73.3128 \pm 1.1746 $ & $ 52.2082 \pm 1.246 $  \\ 
\multicolumn{1}{l|}{}                          & car        & $ 0.9347 \pm 0.004 $  & $ 1.0 \pm 0.0 $       & $ 1.0 \pm 0.0 $ & $ 61.4086 \pm 0.9383 $ & $ 72.6142 \pm 1.3357 $ & $ 51.6545 \pm 1.3383 $ \\ 
\multicolumn{1}{l|}{}                          & chair      & $ 0.926 \pm 0.0045 $  & $ 0.9994 \pm 0.0001 $ & $ 1.0 \pm 0.0 $ & $ 48.9237 \pm 0.9275 $ & $ 63.1441 \pm 1.0583 $ & $ 38.1275 \pm 1.1302 $ \\ \hline
\multicolumn{1}{l|}{\multirow{3}{*}{car}}      & airplane   & $ 0.9279 \pm 0.0027 $ & $ 0.9997 \pm 0.0 $    & $ 1.0 \pm 0.0 $ & $ 50.7472 \pm 1.1176 $ & $ 66.2 \pm 1.0652 $    & $ 40.6049 \pm 1.3128 $ \\ 
\multicolumn{1}{l|}{}                          & car        & $ 0.9585 \pm 0.0025 $ & $ 1.0 \pm 0.0 $       & $ 1.0 \pm 0.0 $ & $ 67.7243 \pm 0.6554 $ & $ 78.7286 \pm 0.797 $  & $ 59.6723 \pm 0.9245 $ \\ 
\multicolumn{1}{l|}{}                          & chair      & $ 0.9218 \pm 0.0035 $ & $ 0.9994 \pm 0.0001 $ & $ 1.0 \pm 0.0 $ & $ 47.7712 \pm 1.1373 $ & $ 63.8324 \pm 1.1691 $ & $ 37.6786 \pm 1.2958 $ \\ \hline
\multicolumn{1}{l|}{\multirow{3}{*}{chair}}    & airplane   & $ 0.9401 \pm 0.0031 $ & $ 0.9998 \pm 0.0 $    & $ 1.0 \pm 0.0 $ & $ 53.7969 \pm 0.8777 $ & $ 67.1421 \pm 1.0913 $ & $ 43.3759 \pm 1.0926 $ \\ 
\multicolumn{1}{l|}{}                          & car        & $ 0.945 \pm 0.0006 $  & $ 1.0 \pm 0.0 $       & $ 1.0 \pm 0.0 $ & $ 63.4508 \pm 0.5761 $ & $ 74.5965 \pm 0.7858 $ & $ 54.3188 \pm 0.8313 $ \\ 
\multicolumn{1}{l|}{}                          & chair      & $ 0.9703 \pm 0.0007 $ & $ 1.0 \pm 0.0 $       & $ 1.0 \pm 0.0 $ & $ 64.3079 \pm 0.4712 $ & $ 75.6459 \pm 0.657 $  & $ 55.2757 \pm 0.7011 $ \\ \hline
\end{tabular}%
}
\end{sc}
\end{small}
\end{center}
\end{table*}

\subsection{DeepEMD as a loss}
\begin{figure*}[!h]
\centering
\hspace*{-2em}
\begin{tabular}{c|cc|cc|cc}
&\multicolumn{2}{c|}{Airplane} & \multicolumn{2}{c|}{Chair} & \multicolumn{2}{c}{Car}\\
\hline 
\rotatebox{90}{\qquad \quad EMD}
&\includegraphics[width=0.15\linewidth,trim=50 45 35 0, clip]{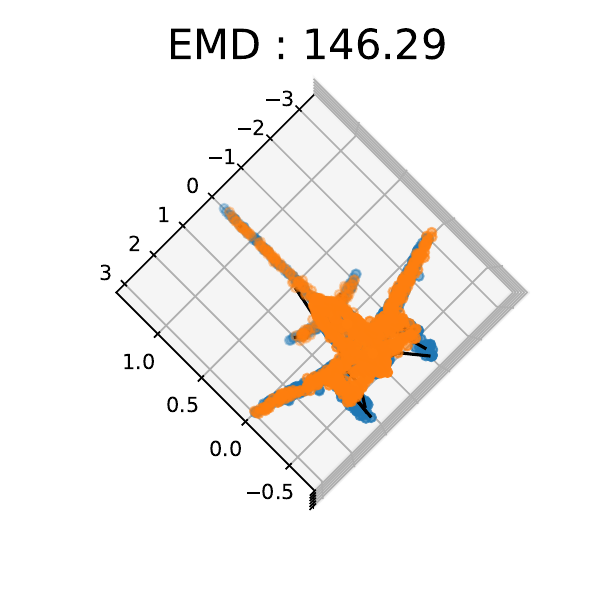}&
\includegraphics[width=0.15\linewidth,trim=50 45 35 10, clip]{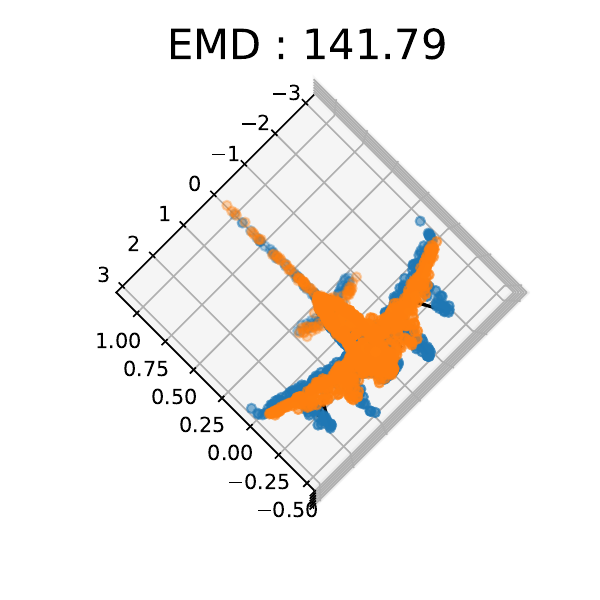}&
\includegraphics[width=0.15\linewidth,trim=50 45 35 10, clip]{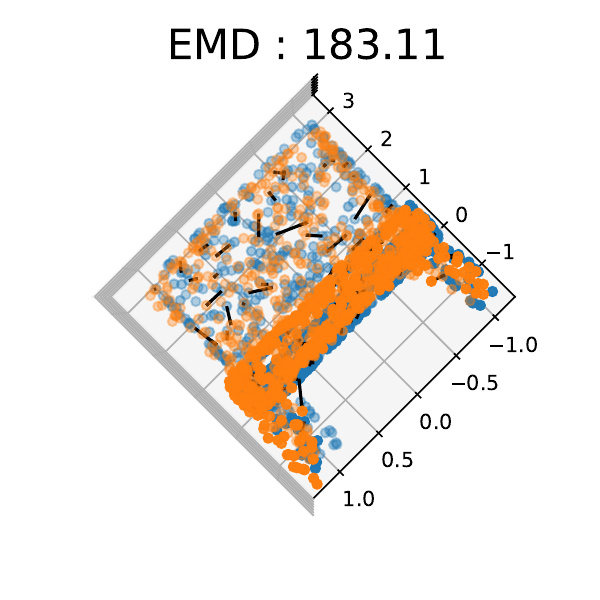}&
\includegraphics[width=0.15\linewidth,trim=50 45 35 10, clip]{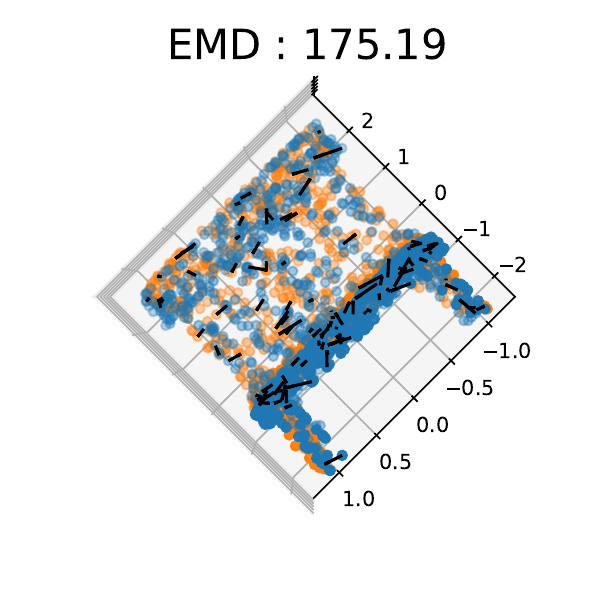}& 
\includegraphics[width=0.15\linewidth,trim=50 45 35 10, clip]{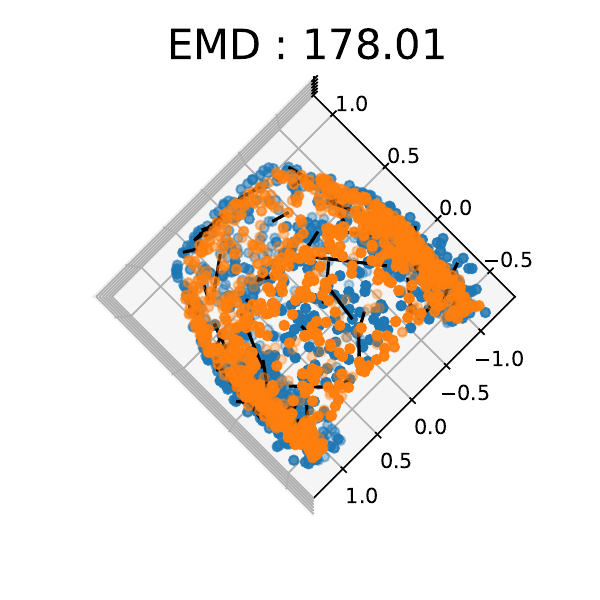}&
\includegraphics[width=0.15\linewidth,trim=50 45 35 10, clip]{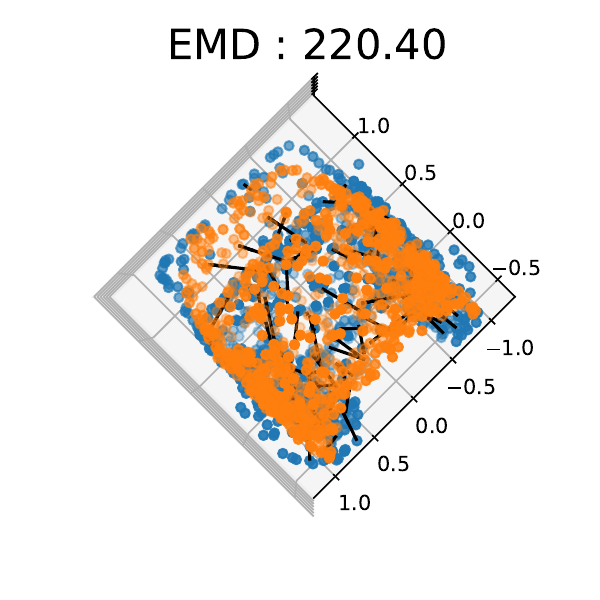} \\[1.5ex]
\rotatebox{90}{\qquad Chamfer}
&\includegraphics[width=0.15\linewidth,trim=50 45 35 0, clip]{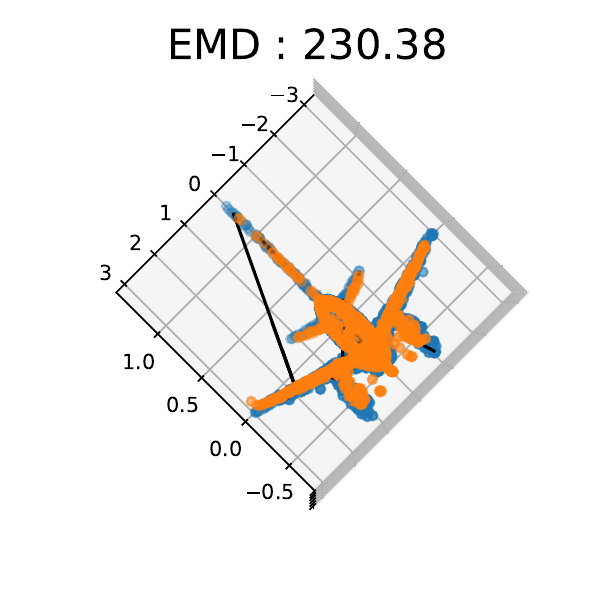}&
\includegraphics[width=0.15\linewidth,trim=50 45 35 10, clip]{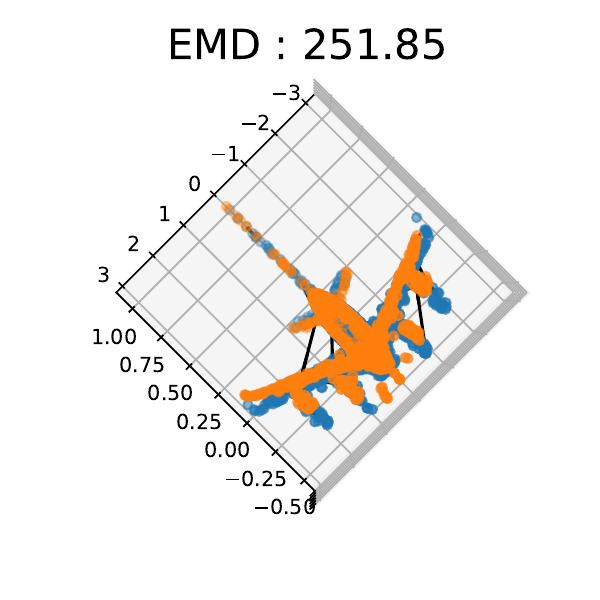}&
\includegraphics[width=0.15\linewidth,trim=50 45 35 10, clip]{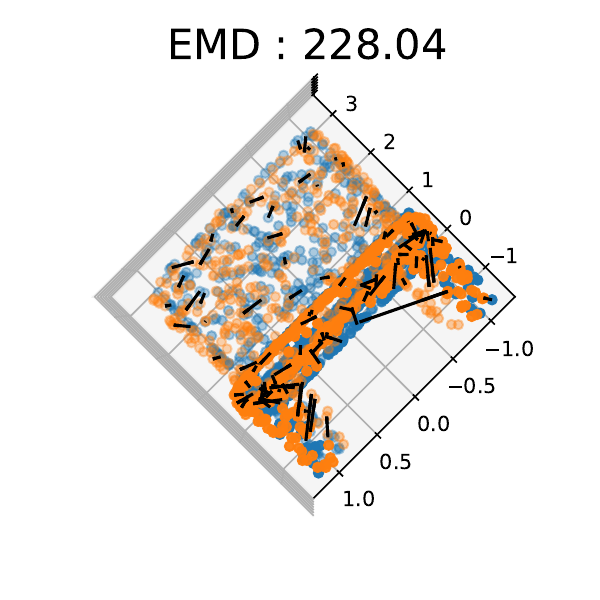}&
\includegraphics[width=0.15\linewidth,trim=50 45 35 10, clip]{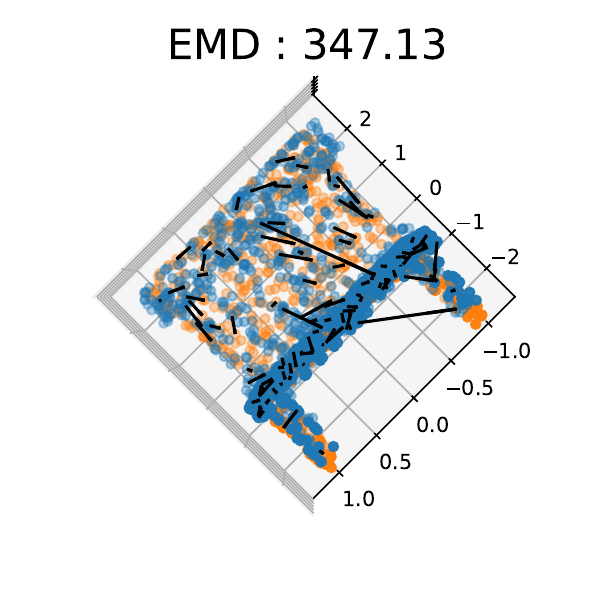}&
\includegraphics[width=0.15\linewidth,trim=50 45 35 10, clip]{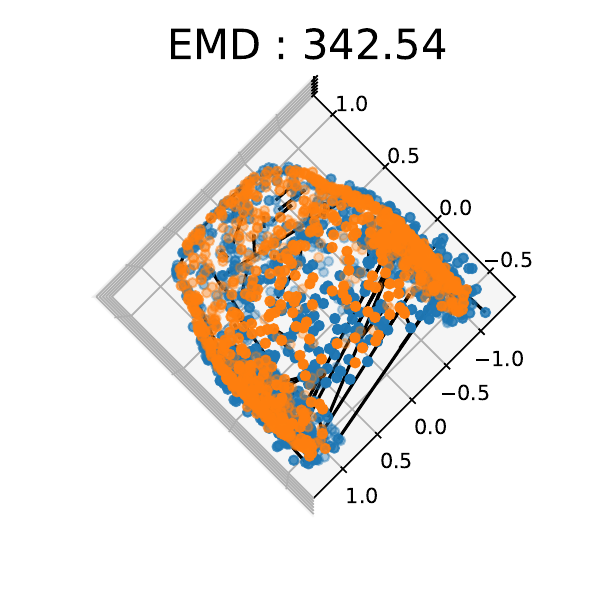}&
\includegraphics[width=0.15\linewidth,trim=50 45 35 10, clip]{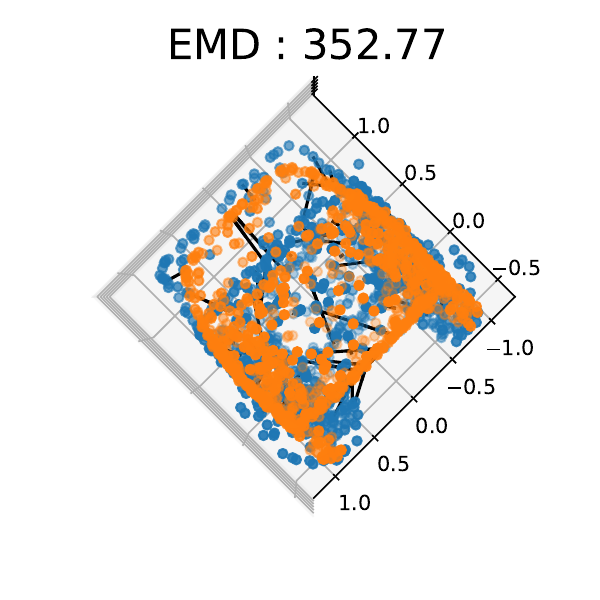} \\[1.5ex]
\rotatebox{90}{\quad DeepEMD (ours)}
&\includegraphics[width=0.15\linewidth,trim=50 45 35 0, clip]{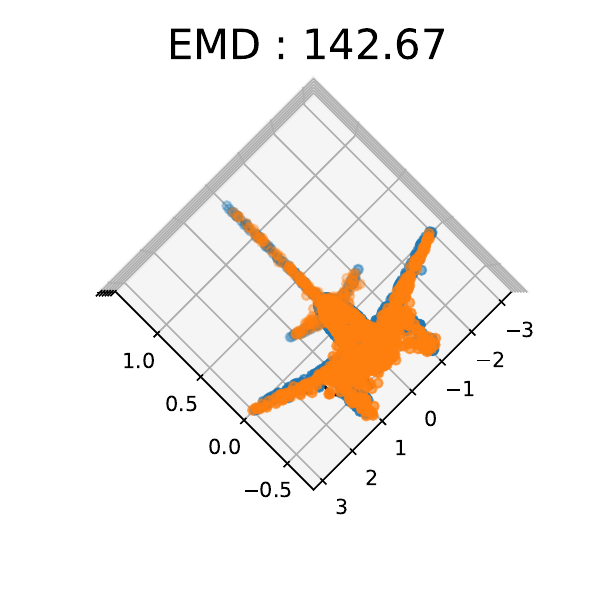}&
\includegraphics[width=0.15\linewidth,trim=50 45 35 10, clip]{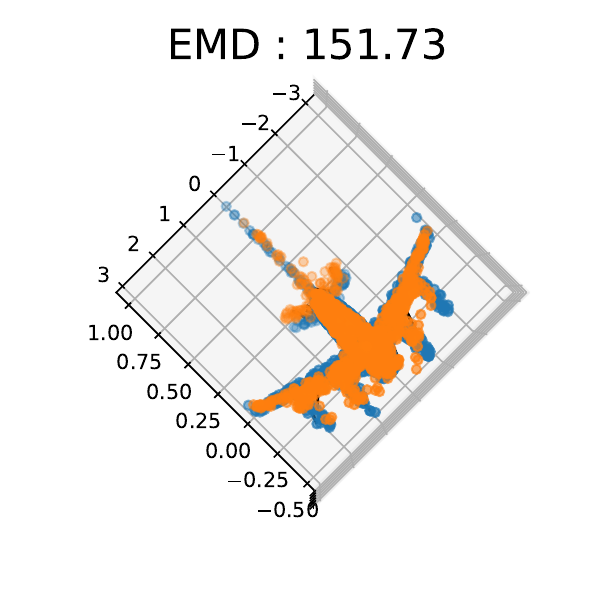}&
\includegraphics[width=0.15\linewidth,trim=50 45 35 10, clip]{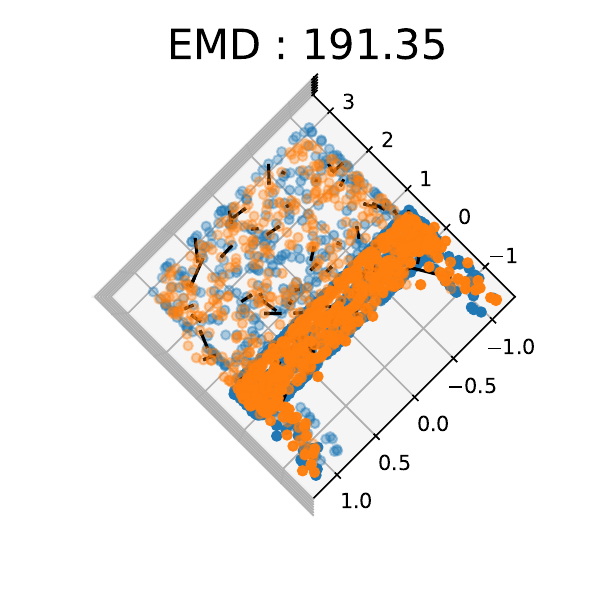}&
\includegraphics[width=0.15\linewidth,trim=50 45 35 10, clip]{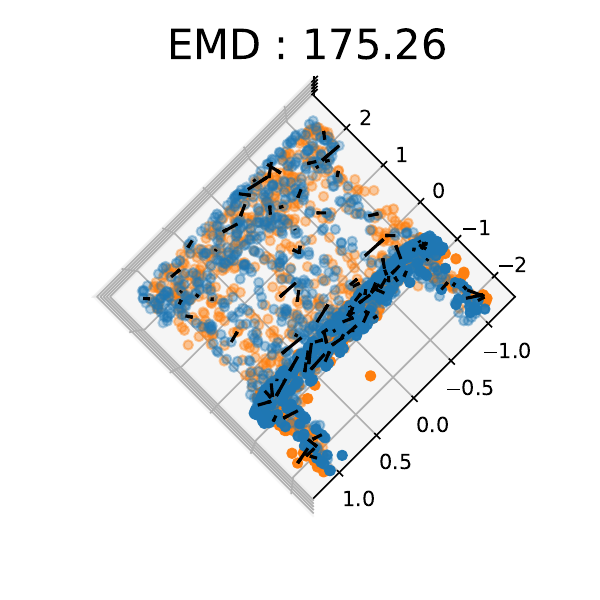}& 
\includegraphics[width=0.15\linewidth,trim=50 45 35 10, clip]{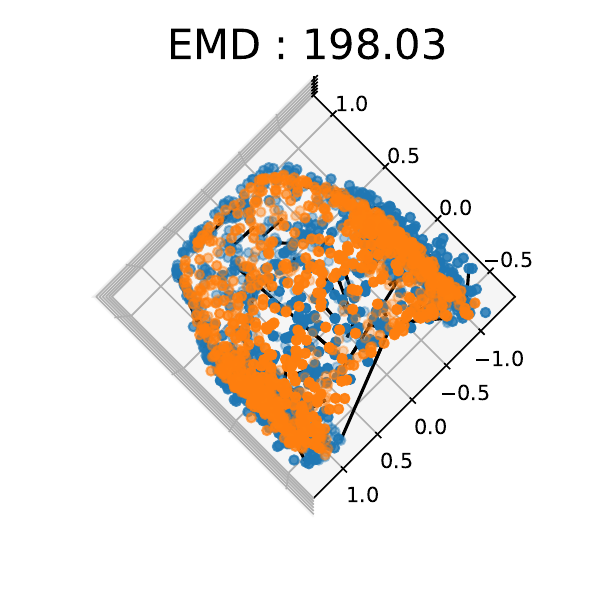}&
\includegraphics[width=0.15\linewidth,trim=50 45 35 10, clip]{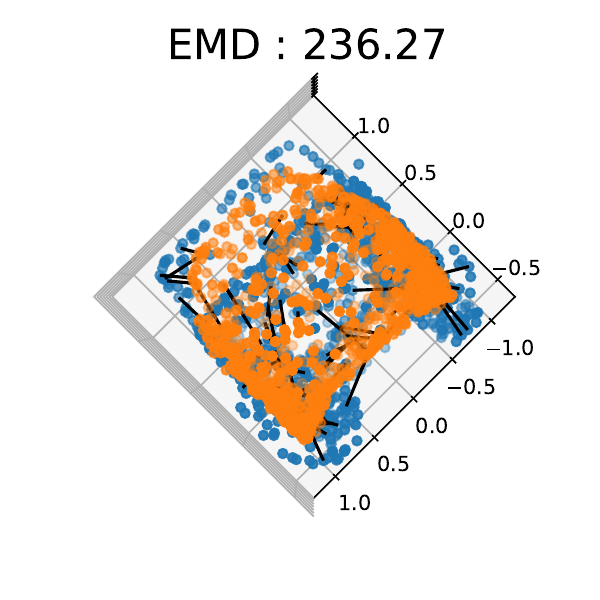}\\
\end{tabular}
\caption{Reconstruction on validation data with SetVAE trained with differnt reconstruction losses : EMD (top), Chamfer (middle) and DeepEMD surrogate (bottom). Training with DeepEMD as a loss consistently achieves lower reconstruction EMD as compared to Chamfer loss.}
\label{fig:val_recons_samples}
\end{figure*}

Fig. \ref{fig:val_recons_samples} shows more samples with the input point cloud and the reconstructed output from SetVAE when trained with EMD, Chamfer or DeepEMD as the reconstruction loss.

\end{document}